%% file: main.tex
\documentclass{article} %
\usepackage{iclr2027_conference,times}

\input{math_commands.tex}

\input{macros}

\usepackage[utf8]{inputenc}
\usepackage[T1]{fontenc}
\usepackage{booktabs}
\usepackage{amsfonts}
\usepackage{microtype}
\usepackage{xcolor}
\usepackage{url}
\usepackage{wrapfig}

\usepackage{amsmath}
\usepackage{graphicx}
\usepackage{float}

\title{DeVI: Physics-based Dexterous Human-Object Interaction via Synthetic Video Imitation}

\author{Hyeonwoo Kim$^1$, Jeonghwan Kim$^1$, Kyungwon Cho$^1$, Junyoung Lee$^1$, Hanbyul Joo$^{1,2}$\\
$^1$Seoul National University, $^2$RLWRLD\\
\texttt{\{hwkim408,roastedpen,cscandkswon,juncong,hbjoo\}@snu.ac.kr}
}

\iclrfinalcopy
\renewcommand{\headrulewidth}{0pt}
\begin{document}

\maketitle

\begin{figure}[H]
    \vspace{-20pt}
    \centering
    \includegraphics[width=\linewidth]{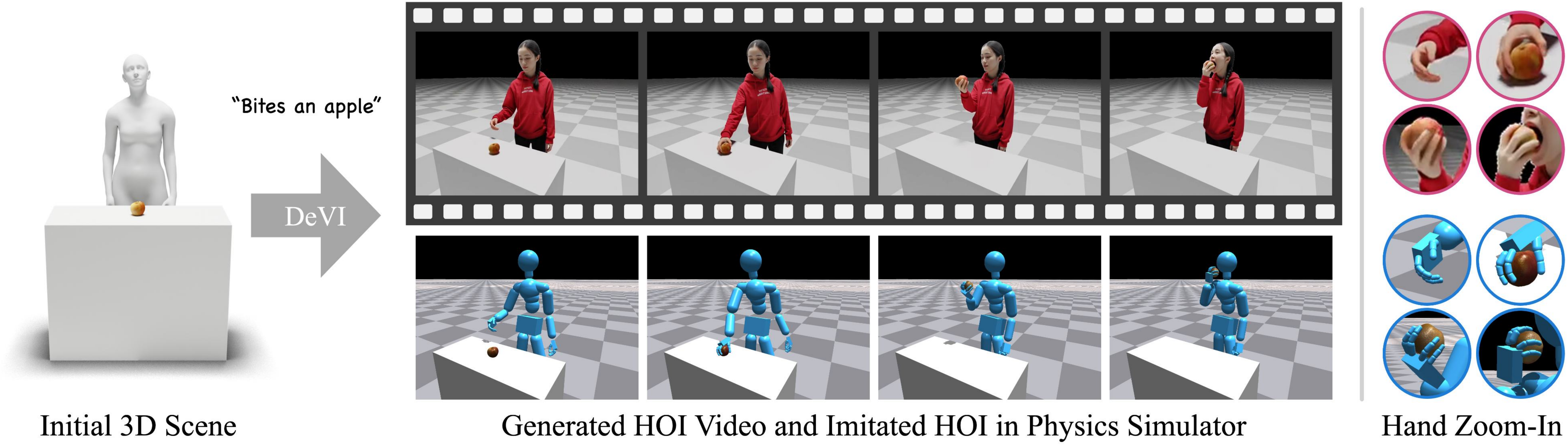}
    \caption{\textbf{DeVI.} Given a physics environment with 3D human and objects along with an interaction text prompt, our method, DeVI, generates a physically plausible dexterous human-object interaction motion by using a video diffusion model as an interaction-aware motion planner.}
    \label{fig:teaser}
\end{figure}

\input{sections/00_abstract}

\input{sections/01_intro}

\input{sections/02_related}

\input{sections/04_method}
\input{sections/05_experiments}

\input{sections/06_discussion}

\bibliography{iclr2027_conference}
\bibliographystyle{iclr2027_conference}

\newpage
\appendix
\input{sections/99_appendix}

\end{document}

%% file: math_commands.tex
\usepackage{amsmath,amsfonts,bm}

\def\eqref#1{equation~\ref{#1}}

\def\1{\bm{1}}

\DeclareMathAlphabet{\mathsfit}{\encodingdefault}{\sfdefault}{m}{sl}
\SetMathAlphabet{\mathsfit}{bold}{\encodingdefault}{\sfdefault}{bx}{n}

%% file: macros.tex
\usepackage{xspace}
\usepackage{bm}
\usepackage{multirow}
\usepackage{gensymb}
\usepackage{tcolorbox}
\usepackage{caption}
\usepackage{capt-of}

\newcommand{\eg}{\textit{e.g.}\xspace}

\renewcommand{\vec}[1]{\bm{#1}}

\usepackage{algorithm}
\usepackage{algorithmic}

\usepackage{hyperref}
\hypersetup{
  colorlinks=true,
  linkcolor=red,     
  urlcolor=magenta,
  citecolor=teal!80!black
}

%% file: sections/00_abstract.tex
\begin{abstract}
Physics-based character control for dexterous Human-Object Interaction (HOI) relies heavily on motion-capture data, which limits its applicability beyond captured scenarios.
Meanwhile, video generative models can synthesize realistic HOI videos for unseen objects and scenarios directly from text prompts, but 3D interactions learned from such synthetic data remain kinematic 
without ensuring physical plausibility.
Imitating these videos with physics-based control is a promising direction, yet unreliable object tracking and hand-object misalignment make it difficult to obtain accurate imitation targets.
We present DeVI (Dexterous Video Imitation), a novel framework that learns physics-based character control for dexterous HOI by imitating text-conditioned synthetic videos.
To address unreliable 3D object motion reconstruction, we introduce hybrid imitation targets that combine reconstructed 3D human motion with tracked 2D object trajectories, bypassing the explicit reconstruction of the 3D object motion.
Our Visual HOI Alignment further refines the human reference to align with both the generated video and the initial 3D object configuration, making it suitable for physical interaction.
Extensive experiments show that DeVI's hybrid representation is more effective than those used by the baselines and that Visual HOI Alignment improves reconstruction quality and imitation success.
We further demonstrate physically plausible functional manipulation from text instructions, including target-aware interactions and articulated object manipulation, without requiring motion-capture demonstrations.
Project page is available at: \color{magenta}{\url{https://snuvclab.github.io/devi/}}
\end{abstract}

%% file: sections/01_intro.tex
\section{Introduction}
\label{sec:intro}

Modeling physics-based human motion is essential for understanding realistic human behavior involving physical interaction with the environment, including dexterous object manipulation.
Existing studies on human motion simulation~\citep{DeepMimic, PHC, PULSE, ADD, MaskedMimic, AMP, ASE, CLoSD, PhysDiff} primarily focus on human motion alone, while the approaches addressing HOI in physical environments often overlook dexterous hand-object interactions involving finger articulation~\citep{AnySkill, GROVE} or focus on specific domains such as sports~\citep{SMPLOlympics, physics_sport_01, physics_sport_02, physics_sport_03}.
Recent approaches show promising results in imitating dexterous HOI~\citep{PhysHOI, HOIfHLI, SimGenHOI} and in learning unified control policies~\citep{SkillMimic, InterMimic}, but still rely on 3D HOI motion-capture (MoCap) demonstrations~\citep{GRAB, OMOMO}, whose limited coverage constrains the diversity of interactions they can learn.
Since dexterous manipulation is difficult to capture at scale, extending physics-based character control to diverse objects and novel functional interactions remains challenging.

Meanwhile, video generative models~\citep{CogVideoX, Wan} enable zero-shot synthesis of high-fidelity HOI videos for unseen objects and scenarios from text prompts, offering rich knowledge of grasping and functional object use (\eg, drinking coke, opening a microwave).
Recent methods leverage such synthetic videos for HOI generation~\citep{DAViD, HOIPAGE}, but the resulting interactions remain kinematic rather than being driven by physics-based control.
Imitating these videos with existing physics-based HOI methods~\citep{InterMimic, PhysHOI, SkillMimic} is a promising direction, but it requires both human and object motion in 3D.
While human motion can be recovered via HMR (Human Mesh Recovery)~\citep{gvhmr, tram}, as in human-only video imitation~\citep{ASAP, physics_for_kinematics_02}, 
3D object motion reconstruction from synthetic videos~\citep{RigVid, Dex4D} remains unreliable due to monocular depth ambiguity.
The resulting trajectory errors cause hand-object misalignment and hinder physics-based imitation (Fig.~\ref{fig:qual2}).
Even with accurate human and object references, universal policies trained on limited 4D HOI datasets fail to generalize to such novel interactions (Fig.~\ref{fig:universal_failure}).

To address these challenges, we present \textbf{DeVI (Dexterous Video Imitation)}, a framework that learns physics-based character control for dexterous HOI by imitating text-conditioned synthetic videos, without requiring 3D HOI MoCap demonstrations.
A key advantage of DeVI is that it learns physics-based control for novel functional interactions, from grasp acquisition to text-driven object use.
To bypass unreliable 3D object motion reconstruction, we introduce \emph{hybrid imitation targets} that combine reconstructed 3D human motion with 2D object tracks from a generated HOI video.
Notably, we observe that precise alignment between the lifted 3D human motion and the initial 3D scene is crucial for
physical interaction (Fig.~\ref{fig:visual_hoi}).
We therefore introduce \emph{Visual HOI Alignment} to refine the human motion to match both the reference video and the initial 3D object configuration.
The resulting hybrid targets are used to train a humanoid control policy via reinforcement learning (RL) with a hybrid tracking reward that combines 3D human and 2D object tracking.

We evaluate DeVI on both generated HOI videos with Internet-sourced objects~\citep{SketchFab} and 3D MoCap demonstrations from GRAB~\citep{GRAB}.
For video-based imitation, we compare our hybrid imitation targets against 
reconstructed 3D object motion~\citep{RigVid, Dex4D, DAViD} and 4D HOI references~\citep{CARI4D, HOIPAGE}.
For MoCap-based imitation, we compare against methods that directly imitate 3D HOI motions~\citep{PhysHOI, SkillMimic, InterMimic}, demonstrating the benefits of our hybrid tracking formulation even when accurate 3D references are available.
We further demonstrate the advantages of synthetic video imitation through diverse functional manipulation scenarios, including target-aware interactions and articulated object manipulation.
Finally, we validate Visual HOI Alignment through ablation studies and comparisons with existing HMR methods~\citep{tram, PromptHMR, gvhmr}.

In summary, our main contributions are as follows.
(1) We introduce DeVI, a framework that learns physics-based character control for dexterous HOI by imitating text-conditioned synthetic videos.
(2) We present hybrid imitation targets that combine reconstructed 3D human motion with 2D object tracks, bypassing explicit 3D object motion reconstruction.
(3) We propose Visual HOI Alignment to refine reconstructed human motion to align with both the reference video and the initial 3D object.
(4) We demonstrate the advantages of synthetic video imitation through various text-driven functional manipulation scenarios, including target-aware interactions and articulated object manipulation.
Our code and results will be publicly released to support reproducibility.

%% file: sections/02_related.tex
\section{Related Work}
\label{sec:related}

\noindent\textbf{Video Generation for Robotic Manipulation.}
Recent advances in video generation motivate the use of generated videos as visual plans for robotic manipulation.
Building on this, recent approaches translate visual predictions into robot control either by learning robot actions from video representations~\citep{UniPi, GR1, VidMan, Vidar} or by recovering explicit geometric motion, including 6D tool trajectories~\citep{Dreamitate} and 6D object trajectories~\citep{RigVid}.
Recent methods extend the paradigm to dexterous manipulation with robotic hands~\citep{VPP, LVP, Dex4D}.
These methods primarily target robot task execution, while physics-based full-body HOI imitation must also include human motion, making it particularly sensitive to spatio-temporal alignment between 3D human motion and object trajectories.
We address the issue through our hybrid imitation targets that combine 3D human motion with 2D object tracks, avoiding explicit 3D object trajectory reconstruction from the video.

\noindent\textbf{Monocular HOI Reconstruction.}
Recovering 3D structure from 2D observations is a long-standing challenge in computer vision due to inherent depth ambiguity.
Recent advances substantially improve monocular 3D reconstruction of scenes~\citep{MonoScene, DepthAnything, DUSt3R}, objects~\citep{InstantMesh, TRELLIS, SAM3D}, and humans~\citep{HMR, gvhmr, tram}.
However, jointly reconstructing a human interacting with an object is considerably more difficult, as it requires precise spatio-temporal alignment between the two.
Early approaches rely on pre-defined object templates and hand-crafted contact heuristics~\citep{PHOSA}.
Subsequent studies replace such heuristics with learning-based contact estimators~\citep{CHORE} and extend the setting to a category-agnostic one~\citep{InteractVLM}, yet remain restricted to single-frame reconstruction.
Recent works extend HOI reconstruction to video-based tracking~\citep{VisTracker, InterTrack, CARI4D}, but remain limited to body-level interactions without modeling dexterous hand motion.
To tackle the challenging nature of 4D HOI reconstruction with dexterous hands, we introduce Visual HOI Alignment, which aligns the reconstructed human motion with the generated video and the object in the initial 3D scene.

\noindent\textbf{Physics-based HOI Motion Generation.}
Traditional studies~\citep{DeepMimic} learn a humanoid control policy via RL to imitate a single MoCap reference motion.
Since learning a separate policy per motion is expensive, several studies learn an integrated policy through adversarial motion priors~\citep{AMP, ASE, CALM} that capture the distribution of motions in the dataset.
To further generalize humanoid control across diverse motions, several approaches~\citep{PHC, PULSE, MaskedMimic} train a unified policy imitating the given demonstrations.
This line of work is extended to HOI, achieving single HOI imitation~\citep{PhysHOI}, interaction skill learning~\citep{PSCI, UniHSI, InterScene, CooHOI, SkillMimic, ResMimic}, and general HOI imitation~\citep{InterMimic}.
However, these approaches rely on 3D human-object motion references that are difficult to obtain at scale, while controllers trained on such MoCap data show limited generalization beyond their training distribution.
A recent study~\citep{ZeroHOI} generates physically plausible HOI motion via a text-to-motion planner, but also shows limited generalization and starts from a pre-grasped state, skipping the important and challenging grasping stage.
We mitigate this problem by using synthetic video as a motion planner that covers the full manipulation sequence, including approach, grasp, and release.

%% file: sections/04_method.tex
\section{DeVI: Dexterous Video Imitation}
\label{sec:method}
Given an initial scene consisting of a 3D human, represented by an SMPL-X~\citep{SMPLX} pose, and a 3D object, our goal is to learn a policy $\pi$ that imitates a text-conditioned synthetic video including dexterous HOI in a physics environment.
Our core idea for imitating the synthetic video is to extract hybrid imitation targets from it, which include 3D human motion and 2D object trajectories, and to use them to train the physics-based character control policy.

In this paper, we focus on dexterous hand-object manipulation in tabletop scenarios.
We initialize a scene $\mathcal{S} = \{\mathcal{H}, \mathcal{O}\}$ as detailed in Appendix~\ref{subsec:scene_init}, where $\mathcal{H}$ denotes the human parameterized by SMPL-X, and $\mathcal{O}$ represents the object defined as follows:
\begin{align}
    \mathcal{H} &= \{ \vec{\beta}, \vec{\theta}^b, \vec{\phi}^b, \vec{\tau}^b, \vec{\theta}^h \}, \quad
    \mathcal{O} = \{ \vec{\phi}^o, \vec{\tau}^o \},
\end{align}
where $\vec{\beta} \in \mathbb{R}^{10}$, $\vec{\theta}^b \in \mathbb{R}^{21\times3}$, $\vec{\theta}^h \in \mathbb{R}^{30\times3}$, $\vec{\phi}^b \in \text{SO(3)}$,  and $\vec{\tau}^b \in \mathbb{R}^{3}$ are the SMPL-X parameters of human shape, body pose, hand pose, global body root orientation, and global body root translation, respectively.
The $\vec{\phi}^o \in \text{SO(3)}$ and $\vec{\tau}^o \in \mathbb{R}^{3}$ are the object's global orientation and  translation.
We define  $\mathcal{M}_{O}(\vec{\phi}^o, \vec{\tau}^o)$ as a mapping from the object state to the transformed object mesh.
We use a similar function $\mathcal{M}_{\text{SMPLX}}(\mathcal{H})$ to obtain the SMPL-X mesh from the parameters.
In this section, we consider the single-object case for simplicity, although our framework readily extends to multi-object.

\begin{figure}[t]\centering
\includegraphics[width=\linewidth, trim={0 0 0 0},clip]{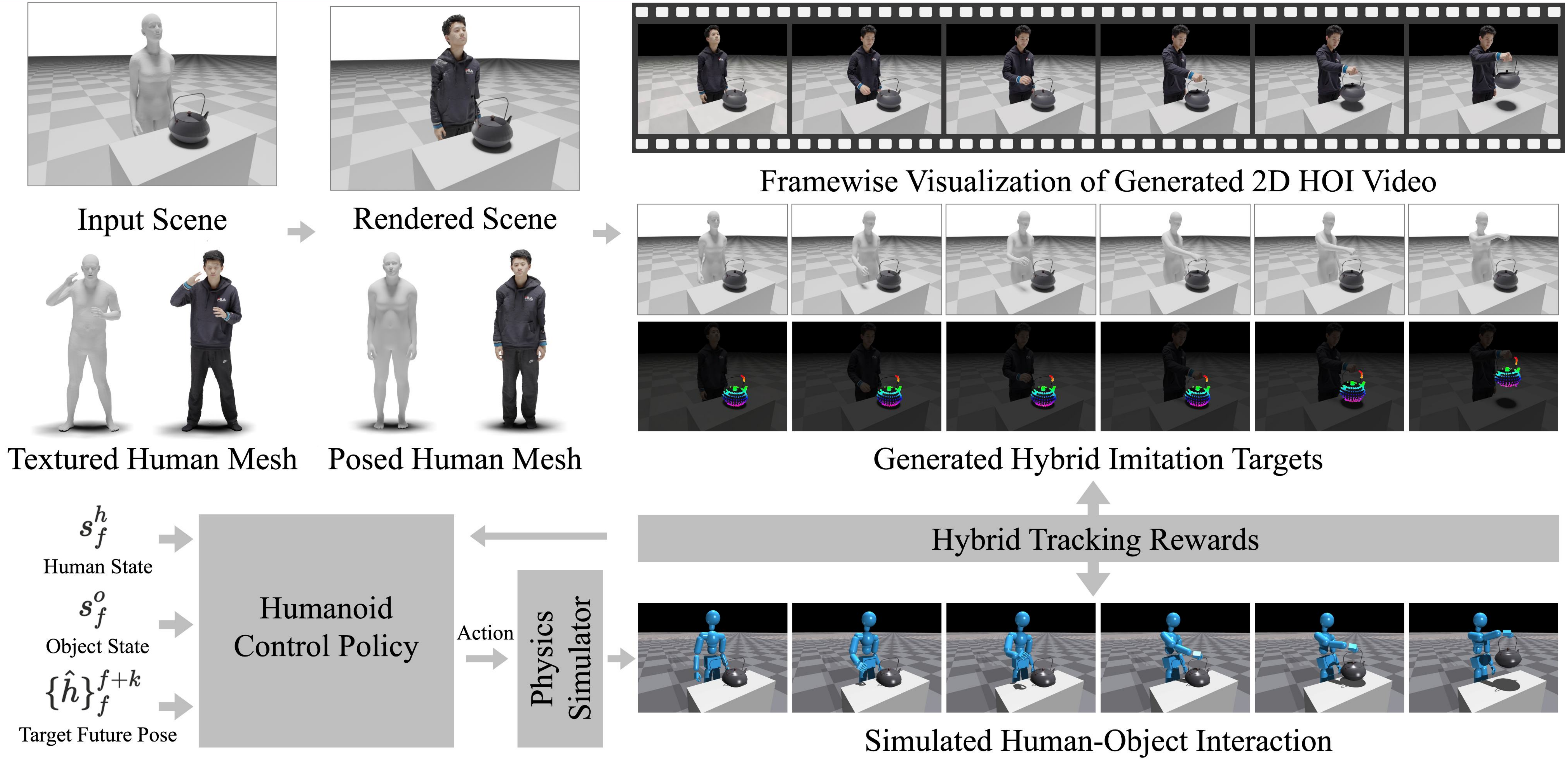}
\caption{\textbf{Overview.} Given a scene with an SMPL-X~\citep{SMPLX} human and object, we replace it with a deformed textured mesh and render to generate HOI video. Then hybrid imitation targets extracted from the video are used to train our humanoid control policy.
}
\label{fig:overview}
\vspace{-5pt}
\end{figure}

We first synthesize an HOI video from the input scene following the prompt (Sec.~\ref{subsec:hoi_video_generation}). From the video, we obtain hybrid imitation targets (Sec.~\ref{subsec:hybrid_imitation_targets}). Then, we present our hybrid tracking reward (Sec.~\ref{subsec:imitate_video}) for policy learning. An overview of our method is shown in Fig.~\ref{fig:overview}.

\subsection{2D HOI Video Generation}
\label{subsec:hoi_video_generation}
As 2D video synthesis and 3D human mesh recovery perform better with realistic appearances, we replace the SMPL-X surface mesh with a textured human mesh $\mathcal{M}_{\text{Human}}(\mathcal{H})$ that matches the scale, pose, and location of $\mathcal{M}_{\text{SMPLX}}(\mathcal{H})$.
In practice, we select textured human meshes from the THuman2.0 dataset~\citep{THuman2.0} and deform them by automatic rigging with approximated joint offsets and skinning weights for linear blend skinning (LBS). 
Appendix~\ref{subsec:rigging} details the rigging process.
We then specify a camera with parameters $\Pi$ and render the scene into 2D image space $\mathcal{I}=\Pi(\{ \mathcal{M}_{\text{Human}}(\mathcal{H}), \mathcal{M}_{O}(\mathcal{O}) \} )$, together with a fixed table and background.
Using a pre-trained image-to-video model~\citep{Wan} and the text prompt, we generate a 2D HOI video $\mathcal{V} = \{\mathcal{I}_t\}_{t=1}^F$ initialized from the rendered image $\mathcal{I}_1 = \mathcal{I}$.

\subsection{Extracting Hybrid Imitation Targets}
\label{subsec:hybrid_imitation_targets}
Directly using the generated video $\mathcal{V}$ for RL policy learning is challenging.
Instead, we first extract hybrid imitation targets $\hat{\vec{g}}^{\text{hybrid}} = \{ \hat{\vec{h}}, \hat{\vec{o}} \}$, which consist of estimated 3D human motion $\hat{\vec{h}}$ and 2D object trajectories $\hat{\vec{o}}$ from the video as follows:
\begin{align}
    \hat{\vec{h}} &= \{ \hat{\vec{J}}_t^b, \hat{\vec{\theta}}_t^b, \hat{\vec{J}}_t^h, \hat{\vec{\theta}}_t^h \}_{t=1}^F, \quad
    \hat{\vec{o}} = \{ \hat{\vec{x}}_t \}_{t=1}^F.
\end{align}
Here, $\hat{\vec{J}}_t^b \in \mathbb{R}^{19 \times 3}$ and $\hat{\vec{J}}_t^h \in \mathbb{R}^{32 \times 3}$ denote the estimated 3D body and hand joint locations, and $\hat{\vec{\theta}}_t^b \in \mathbb{R}^{19 \times 3}$ and $\hat{\vec{\theta}}_t^h \in \mathbb{R}^{32 \times 3}$ represent the corresponding body and hand pose parameters in SMPL-X representation.
The object target $\hat{\vec{o}}$ consists of tracked 2D points $\hat{\vec{x}}_t \in \mathbb{R}^{M \times 2}$ in image coordinates at time $t$, initialized from the first frame. These points correspond to $M$ visible object vertices projected into the image using the camera $\Pi$.
Our motivation for using 3D human pose with 2D object trajectories is that accurately recovering full 3D object pose from video remains challenging as shown in Fig.~\ref{fig:qual2}, and robust 3D spatial alignment between humans and objects is still an open problem. By providing these hybrid cues as reference targets, we allow the RL framework to infer a physically plausible solution that jointly imitates both signals.
Below, we describe the extraction process for each component of the hybrid imitation target.

\noindent\textbf{Initializing 3D Human Reference.}
For the human component, we apply an off-the-shelf monocular world-grounded human motion estimator~\citep{gvhmr}, denoted as $\mathcal{F}^b$, along with a hand pose estimator~\citep{HaMeR}, denoted as $\mathcal{F}^h$, to the synthesized 2D video $\mathcal{V}$. These models provide estimates of 3D body motion and 3D hand poses, which we combine to reconstruct a unified SMPL-X human motion sequence with hand poses. Specifically,
\begin{align}
    \mathcal{F}^b(\mathcal{I}_t) &= \{ \vec{\beta}_t, \vec{\theta}^b_t, \vec{\phi}^{b}_t, \vec{\tau}^b_t \}, \quad
    \mathcal{F}^h(\mathcal{I}_t) = \{ \vec{\theta}^h_t, {\vec{\phi}}^{lh}_t, {\vec{\phi}}^{rh}_t, \vec{\tau}^{lh}_t, \vec{\tau}^{rh}_t \},
\end{align}
where $\vec{\beta}_t \in \mathbb{R}^{10}$, $\vec{\theta}^b_t \in \mathbb{R}^{21 \times 3}$, $\vec{\phi}^{b}_t \in \text{SO(3)}$ , and $\vec{\tau}^b_t \in \mathbb{R}^3$ are the estimated SMPL-X parameters for human shape, body pose, the root joint orientations in the world coordinate, and translation, respectively.
The $\vec{\theta}^h_t \in \mathbb{R}^{30 \times 3}$ is a hand pose, ${\vec{\phi}}^{lh}_t, {\vec{\phi}}^{rh}_t \in \text{SO(3)}$ are the root (wrist) orientations of each hand, and $\vec{\tau}^{lh}_t, \vec{\tau}^{rh}_t \in \mathbb{R}^3$ are the global translations of each hand.
As the hand estimator typically provides more accurate hand locations and orientations, we refine the body pose by adjusting the wrist joint angles in $\vec{\theta}^b_t$ to combine the outputs of $\mathcal{F}^b$ and $\mathcal{F}^h$.
Appendix~\ref{subsec:unify_body_hand} details the unifying process of body and hand.
The resulting unified 3D human representation at time $t$ is:
\begin{equation}
    \mathcal{H}_t = \{ \vec{\beta}_t, \vec{\tilde{\theta}}^{b}_t,
    \vec{\phi}^{b}_t, \vec{\tau}^b_t, \vec{\theta}^{h}_t \},
\end{equation}
with adjusted body pose $\vec{\tilde{\theta}}^{b}_t$.
Ideally, the reconstructed SMPL-X human model at the first frame $\mathcal{H}_{t=1}$ should match the initial SMPL-X model $\mathcal{H}$ in the input scene $\mathcal{S}$.
We therefore apply a global rigid transformation to match the position and orientation of $\mathcal{H}_{t=1}$ to the initial state. This transformation is analytically derived from their relative pose, and applied to all subsequent frames, resulting in:
\begin{equation}
    \mathcal{\tilde{H}}_t = \{ \vec{\beta}_t, \vec{\tilde{\theta}}^{b}_t,
    \vec{\tilde{\phi}}^{b}_t, \vec{\tilde{\tau}}^b_t, \vec{\theta}^{h}_t \},
\end{equation}
where $\vec{\tilde{\phi}}^b_t$ and $\vec{\tilde{\tau}}^b_t$ are the adjusted global orientation and translation.

\noindent\textbf{Refining 3D Human Reference via Visual HOI Alignment.}
As the coarse SMPL-X reconstructed in the previous stage is obtained by unifying the outputs of two independent estimators, it is not perfectly aligned with the reference video $\mathcal{V}$ or the 3D object $\mathcal{O}$ in the scene $\mathcal{S}$, as shown in Fig.~\ref{fig:visual_hoi}.

To address this, we present Visual HOI Alignment, which optimizes the body and hand pose parameters to align with both the reference video $\mathcal{V}$ and the initial 3D object state $\mathcal{O}$.
We optimize body pose parameters $\{ \vec{\hat{\theta}}^{b}_t \}_{t=1}^F$ and hand parameters $\{ \vec{\hat{\theta}}^{h}_t \}_{t=1}^F$ by minimizing the following objective:
\begin{align}
    \mathcal{L}_{\text{total}} = w_b\mathcal{L}_b + w_h\mathcal{L}_h + w_{tc}\mathcal{L}_{tc} + w_{\text{HOI}}\mathcal{L}_{\text{HOI}},
\end{align}
where $\mathcal{L}_b$, $\mathcal{L}_h$, $\mathcal{L}_{tc}$, $\mathcal{L}_{\text{HOI}}$ are the
2D body projection loss, 2D hand projection loss, temporal consistency loss, and HOI loss, with their corresponding weights, $w_b$, $w_h$, $w_{tc}$, and $w_{\text{HOI}}$.
The 2D body projection loss $\mathcal{L}_b$ and 2D hand projection loss $\mathcal{L}_h$ are defined to align the projection of SMPL-X joints to the original joint estimations from the $\mathcal{F}^b$ and $\mathcal{F}^h$ in 2D as follows:
\begin{align}
    \mathcal{L}_b &=  \sum_{t=1}^F\| \Pi \left(\mathcal{J}_b( \mathcal{H}_t ) \right) - \vec{j}_t^{\text{body}} \|^2, \quad
    \mathcal{L}_h =  \sum_{t=1}^F\| \Pi \left(\mathcal{J}_h( \mathcal{H}_t ) \right) - \vec{j}_t^{\text{hand}} \|^2,
\end{align}
where $\mathcal{J}_b$, $\mathcal{J}_h$ are the body and hand joint regressors from SMPL-X parameters in $\mathcal{H}_t$. The $ \vec{j}_t^{\text{body}}$ and $\vec{j}_t^{\text{hand}}$ are estimated 2D body and hand joints obtained from $F_b$ and $F_h$.
The temporal consistency loss $\mathcal{L}_{tc}$ encourages temporal consistency, and is defined as:
\begin{align}
    \mathcal{L}_{tc} &= \sum_{t=1}^{F-1} \mathcal{D}_{\text{geo}} (  {\vec{\theta}}^b_t, {\vec{\theta}}^b_{t+1} ) + \sum_{t=1}^{F-1} \mathcal{D}_{\text{geo}} (  {\vec{\theta}}^h_t, {\vec{\theta}}^h_{t+1} ),
\end{align}
where $\mathcal{D}_{\text{geo}}(\cdot)$ is a mean geodesic distance between rotations.
Additionally, we present an HOI loss $\mathcal{L}_{\text{HOI}}$ to enforce the human body parts to be in contact with the object at a certain time instance:
\begin{align}
    \mathcal{L}_{\text{HOI}} &=  \min_t \mathcal{D}_{\text{chamfer}} ( \mathcal{J}_* (v^{\text{SMPLX}}_t), v_* )
\end{align}
where $\mathcal{D}_{\text{chamfer}}(A, B)$ is a one-sided Chamfer~\citep{Chamfer} distance from $A$ to $B$, $\mathcal{J}_*$ is the SMPL-X joint regressor for specific body parts (\eg, left hand), $v^{\text{SMPLX}}_t$ is the SMPL-X vertices at time $t$, and $v_* \in \mathbb{R}^{n_* \times 3}$ is the initialized object vertices. Note that the pair $( \mathcal{J}_*, v_*)$ is specified in the text prompt we used for generating video (\eg, ``holds the \underline{coke} with \underline{left hand}''). The HOI loss is motivated by the intuition that at least one frame of the human motion should establish contact with the initial object to move the object.
The resulting SMPL-X joint positions and rotations after our Visual HOI Alignment construct our 3D human reference $\hat{h}$.

\noindent\textbf{Generating 2D Object Reference.}
For the object side, we first identify the visible vertices and their corresponding projections for the object mesh via ray casting~\citep{raycast} using the camera $\Pi$.
The projected vertices are tracked across video frames using a video tracker~\citep{cotracker3}, constructing our 2D object reference $\hat{\vec{o}} = \{ \hat{\vec{x}}_t \}_{t=1}^F$.
We filter out vertices that are heavily occluded across video frames using the occlusion mask estimated by the tracker.

\subsection{Learning Humanoid Control Policy}
\label{subsec:imitate_video}
We formulate physics-based character control as a Markov Decision Process and optimize the policy with PPO~\citep{PPO}.
Following MaskedMimic~\citep{MaskedMimic}, the human state $\vec{s}_t^h \in \mathbb{R}^{778}$ consists of joint positions, rotations, and linear/angular velocities. The object state $\vec{s}_t^o \in \mathbb{R}^{15}$ includes its position, orientation, and velocities. The action $\vec{a}_t \in \mathbb{R}^{51 \times 3}$ defines PD target for all body and hand joints.
To train humanoid control policy $\pi_\theta(\vec{a}_t | \vec{s}_t, \vec{g}_t)$, we define $\vec{g}_t =\hat{\vec{h}}_{t}^{t+k}$, the future $k$ entries of $\hat{\vec{h}}$ as a goal, while guiding to match $\hat{\vec{o}}$ using our hybrid tracking reward.

\noindent\textbf{Hybrid Tracking Reward.}
Our hybrid tracking reward tracks the imitation targets from the video:
\begin{align}
R(\vec{s}_t, \vec{a}_t, \vec{g}_t) =
   R_h \cdot R_o \cdot R_{\text{contact}},
\end{align}
where $R_h$,  $R_o$, and $R_{\text{contact}}$ are the human tracking, object tracking, and contact rewards.
The human tracking reward $R_h$ encourages the humanoid character to track the synthetic 3D human motion $\hat{\vec{h}}$. It is defined as the product of several joint difference rewards and a power penalty as follows:
\begin{align}
   R_h = r_{\text{jp}} \cdot r_{\text{jv}} \cdot r_{\text{jr}} \cdot r_{\text{lp}}^h \cdot r_{\text{lr}}^h \cdot r_{\text{pw}},
\end{align}
where $r_{\text{jp}}$, $r_{\text{jv}}$, $r_{\text{jr}}$ are the full body joint position, velocity, and rotation rewards.
The terms $r_{\text{lp}}^h$ and $r_{\text{lr}}^h$ are the local hand joint position and rotation reward. These local joint quantities are represented in the wrist-centric coordinate by transforming joint positions and rotations relative to the wrist. The $r_{\text{pw}}$ is a power penalty reward~\citep{QuestEnvSim} preventing excessive forces and guiding smooth motion.
Our rewards are formulated as an exponential of the mean distance between the reference and the simulated state. 
The object tracking reward $R_o$ drives the 2D projection of the simulated object toward the reference trajectory $\hat{\vec{o}} = \{ \hat{\vec{x}}_t \}_{t=1}^F$, defined as:
\begin{align}
   R_o = e^{\lambda_o \| \hat{\vec{x}}_t - \vec{x}_t \|^2},
\end{align}
where $t$ denotes the simulation time step, $\lambda_o$ is a weight coefficient, and $\vec{x}_t \in \mathbb{R}^{M \times 2}$ is the projection of the same $M$ vertices of the simulated object into the image space via the same view with $\Pi$.
In practice, the object starts from its initial 3D state $\mathcal{O}$ in simulation and moves only through physical contact, guided by our object reward.
We detail our reward including coefficients and formulation with contact reward in Appendix~\ref{subsec:hybrid_reward_details} and ~\ref{subsec:contact_reward_details}, and network architecture in Appendix~\ref{subsec:architecture}.

%% file: sections/05_experiments.tex
\section{Experiments}
\label{sec:experiments}

\subsection{Experimental Setup}
\label{subsec:setup}

\noindent\textbf{Imitating Generated Videos.}
We evaluate DeVI on 10 generated videos, each containing a different object and a different interaction.
As no prior physics-based character control method imitates dexterous HOI directly from a generated video, we compare our imitation targets rather than complete systems.
Specifically, we replace our 2D object tracks with the 3D object motion recovered by RIGVid~\citep{RigVid}, DAViD~\citep{DAViD}, and Dex4D~\citep{Dex4D} from the same video,  using the corresponding object tracking rewards.
All settings share the same human reference refined by Visual HOI Alignment and all other rewards, with object rewards and termination criteria matched to each object reference.
As the generated video provides no ground-truth 3D motion, we treat the reconstructed 3D human motion as a pseudo ground truth and report MPJPE (Mean Per Joint Position Error) for body, hand, and all joints.
For the object side, we compare the mask of the simulated object against the video mask obtained from SAM~3.1~\citep{SAM3}, reporting pixel-level IoU, precision, recall, and F1 score.

\noindent\textbf{Imitating MoCap Demonstrations.}
While our goal is to imitate synthetic videos, we additionally compare with methods that imitate 3D MoCap HOI demonstrations in physics simulators.
Specifically, we compare DeVI against state-of-the-art 3D HOI imitation studies,  PhysHOI~\citep{PhysHOI}, SkillMimic~\citep{SkillMimic}, and InterMimic~\citep{InterMimic}, on the GRAB~\citep{GRAB} dataset.
Since DeVI uses 2D object trajectories rather than the 6D poses used by the baselines, we project 1024 vertices sampled from the object surface through a virtual camera and use their per-frame projections as the imitation target.
We train all baselines under the same single-motion setting on 30 HOI motions from the GRAB dataset, and report human MPJPE separately for body, hand, and all joints along with the root joint error $T_{\text{root}}$.
For the object side, we additionally follow CHOIS~\citep{CHOIS} and report metrics for the object translation $T_{\text{obj}}$ and orientation $O_{\text{obj}}$.

\subsection{Synthetic HOI Video Imitation}
\label{subsec:video_imitation}

\noindent\textbf{Functional Manipulation for Various Objects.}
By leveraging a video diffusion model as an HOI-aware motion planner, DeVI allows planning various HOIs from text prompts.
A generated video shows not only how a person grasps an object but also how the person uses the object afterward, providing a motion plan for functional manipulation.
As shown in Fig.~\ref{fig:qual}, DeVI simulates both single-handed interactions (\eg, tossing a wok) and two-handed interactions (\eg, taking photos).
The planner also uses the input scene to guide target-aware placement, such as placing a pan on an induction stove, without an explicit scene understanding module.
With the input scene fixed, different text prompts lead to placing either an apple or a tomato in the basket, demonstrating DeVI's text controllability.
Figs.~\ref{fig:P1} and~\ref{fig:P2} in the appendix detail the video-motion alignment of our policy.

\input{contents/01_main_qual}
\input{contents/03_quant_video_imitation}

\noindent\textbf{Comparison with Baselines.}
For video imitation, we treat a clip as successful when its average all-joint MPJPE stays below $0.2 \text{m}$ and its average mask IoU exceeds $0.5$, and we apply the same criteria per frame to obtain the frame success rate.
As shown in Tab.~\ref{tab:quant_video}, DeVI outperforms the baselines on all metrics.
The baselines rely on explicit 3D object motion reconstruction, which remains unreliable due to monocular depth ambiguity and leads to severe hand-object misalignment (Fig.~\ref{fig:qual2}).
In contrast, DeVI keeps the object target in 2D, allowing policy learning in simulation to find a physically feasible 3D object trajectory consistent with the 2D tracks.
DeVI also outperforms policies trained with 4D HOI references from CARI4D and HOI-PAGE, whose reconstructions exhibit inaccurate object motion or insufficient finger articulation (Appendix Figs.~\ref{fig:qual_4d_hoi} and~\ref{fig:qual_4d_hoi_failure}, Tab.~\ref{tab:quant_4d_hoi}).

\noindent\textbf{Representation Choice.}
We measure the reliability of each representation on 100 GRAB~\citep{GRAB} sequences selected alphabetically and rendered from a fixed viewpoint, providing exact ground-truth object motion.
All baselines receive the same initial 3D scene information as DeVI.
We report vertex, position, and rotation errors, using mean circular difference~\citep{ImageRotation} for image-space rotation error.
DAViD achieves lower 3D rotation error than RIGVid's direct 6D pose estimation and Dex4D's depth lifting, highlighting the strong rotation cues from 2D tracks anchored to known geometry (Tab.~\ref{tab:representation_choice}).
However, its Perspective-n-Point (PnP) reconstruction still yields a position error of $0.2763$~m, too large for dexterous manipulation.
DeVI therefore keeps the tracks in 2D, preserving these rotation cues without relying on reconstructed depth.

\input{contents/02_qual_video_imitation}
\input{contents/04_quant_mocap}

\noindent\textbf{Applications.}
Using generated videos as demonstrations, DeVI extends to egocentric settings, enabling articulated object manipulation with a simulated Inspire Hand RH56DFTP (Fig.~\ref{fig:application}).
We further use hybrid targets from generated egocentric videos to optimize trajectories for real-world manipulation with a robotic arm and dexterous hand (Appendix~\ref{subsec:application_details}).

\subsection{HOI MoCap Imitation}
\label{subsec:mocap_imitation}

\noindent\textbf{Imitation Accuracy.}
As the baselines use different early-termination conditions, the tracking errors of failed rollouts start showing poor accuracy at different frames and are not fairly comparable across methods.
We therefore retain motions where both DeVI and the paired baseline does not collapse.
Specifically, we compare motions with $\text{MPJPE (All)} < 0.2\,\text{m}$ and $T_{\text{obj}} < 0.2\,\text{m}$ to measure the general imitation accuracy with largely overlapped sequences.
As shown in Tab.~\ref{tab:quant_mocap}, DeVI outperforms each baseline on all metrics, demonstrating the benefits of our reward setup even when accurate 3D references are available.
Notably, DeVI with our 2D tracking reward also outperforms a variant trained with the conventional 6D pose tracking reward, showing that the 2D trajectory, which is easier to obtain than 6D pose, can lead to stronger HOI imitation performance.
We attribute this to our implicit 2D supervision, which guides object position and rotation jointly through projected geometry and avoids over-constraining the object trajectory during policy learning.

\noindent\textbf{Imitation Success.}
To evaluate precise imitation for dexterous HOI, we define success with two different thresholds, which share the conditions $\text{MPJPE (All)} < 0.1\,\text{m}$ and $T_{\text{obj}} < 0.1\,\text{m}$ but differ in object orientation $O_{\text{obj}}$.
As shown in Tab.~\ref{tab:quant_mocap_success}, DeVI achieves higher success rates, demonstrating that hybrid targets make feasible interactions easier to learn than conventional 6D pose targets.

\subsection{Visual HOI Alignment}
\label{subsec:visual_hoi_eval}

\noindent\textbf{Ablation Study.}
To verify the efficacy of our Visual HOI Alignment, we conduct qualitative and quantitative ablations.
As shown in Fig.~\ref{fig:visual_hoi}, merging independently estimated body and hand poses into a single SMPL-X~\citep{SMPLX} model leaves severe misalignment with the video and the 3D object, especially around hand-object interactions.
In contrast, our Visual HOI Alignment aligns the motion with both the video and the 3D object, making the reference feasible for physical interaction.
This improvement is also reflected in Tab.~\ref{tab:visual_hoi_ablation}, where Visual HOI Alignment substantially improves imitation accuracy and success rates.
Notably, while the comparison in Tab.~\ref{tab:quant_video} uses a shared 3D human reference, this ablation uses different human references.
We therefore measure image-space MPJPE between 2D keypoints extracted from the video and the projected simulated joints.

\noindent\textbf{Comparison with Human Mesh Recovery.}
For a quantitative evaluation, we compare our reconstructed human motion with existing human mesh recovery methods on the grab sequences of Subject 1 in ARCTIC~\citep{ARCTIC}, which provides RGB images and ground-truth human motion.
As shown in Tab.~\ref{tab:hmr_arctic}, DeVI with Visual HOI Alignment outperforms baselines for all metrics.
Notably, merging independent body and hand estimates alone does not improve hand accuracy, whereas Visual HOI Alignment substantially improves it.

\input{contents/06_visual_hoi_ablation_application}
\input{contents/05_visual_hoi}

%% file: contents/01_main_qual.tex
\begin{figure}[t]\centering
\includegraphics[width=\linewidth, trim={0 0 0 0},clip]{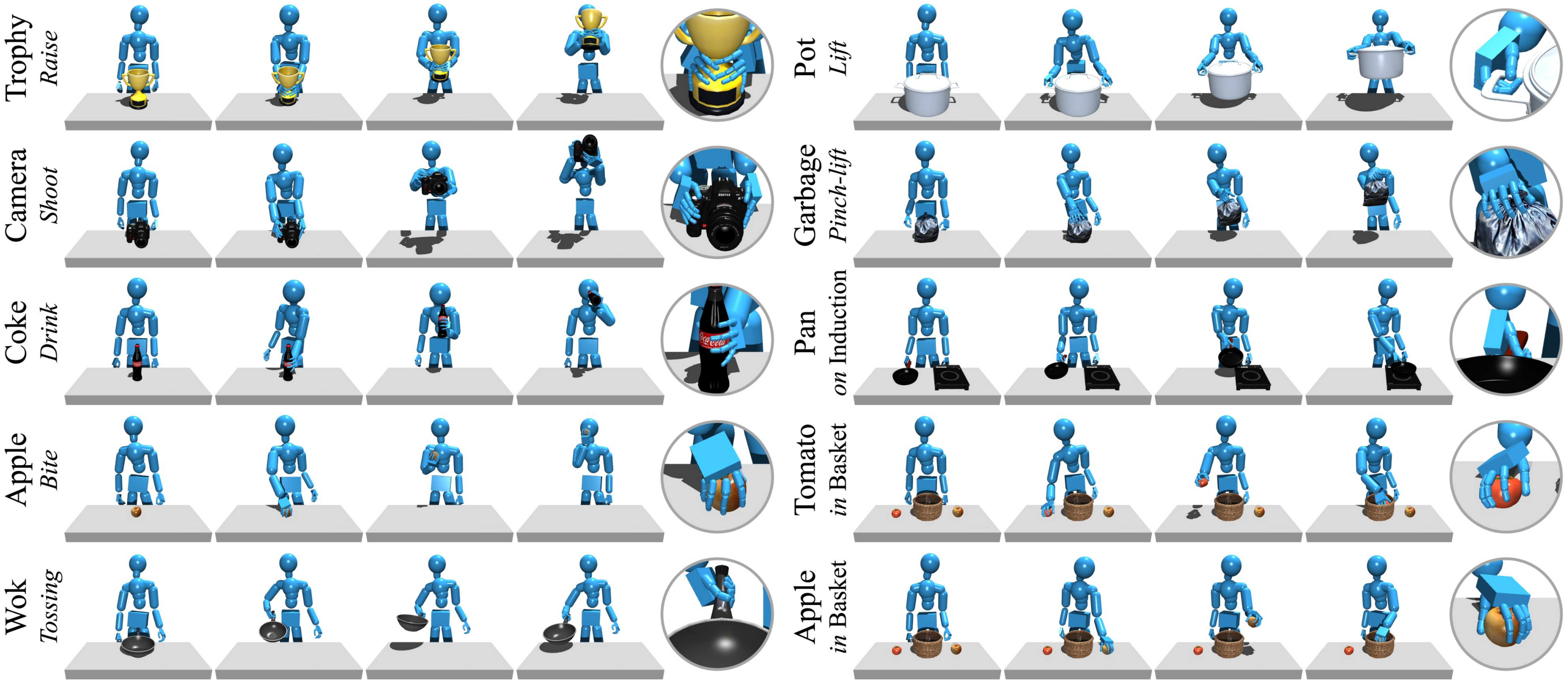}
\caption{\textbf{Qualitative Results on Various Objects.} DeVI leverages a video diffusion model as an HOI-aware motion planner, allowing simulation of HOI with diverse objects through text prompts.
}
\label{fig:qual}
\vspace{-8pt}
\end{figure}

%% file: contents/03_quant_video_imitation.tex
\begin{table}[!t]
\centering
\begin{minipage}[t]{0.54\linewidth}
    \vspace{0pt}
    \centering
    \scriptsize
    \setlength{\tabcolsep}{1.5pt}
    \renewcommand{\arraystretch}{1.25}
    \begin{tabular}{@{}lccccccccc@{}}
        \toprule
        \multirow{2}{*}{\raisebox{-2.0pt}{Methods}} &
        \multicolumn{3}{c}{MPJPE (mm) $\downarrow$} &
        \multirow{2}{*}{\raisebox{-12.0pt}{\shortstack{IoU\\(\%) $\uparrow$}}} &
        \multirow{2}{*}{\raisebox{-12.0pt}{\shortstack{Precision\\(\%) $\uparrow$}}} &
        \multirow{2}{*}{\raisebox{-12.0pt}{\shortstack{Recall\\(\%) $\uparrow$}}} &
        \multirow{2}{*}{\raisebox{-12.0pt}{\shortstack{F1\\(\%) $\uparrow$}}} &
        \multicolumn{2}{c}{Success (\%) $\uparrow$} \\
        \cmidrule(lr){2-4}
        \cmidrule(lr){9-10}
        & Body & Hand & All & & & & & Frame & Clip \\
        \midrule
        RIGVid
        & 641.4 & 684.3 & 666.1
        & 38.03 & 45.70 & 43.43 & 44.29
        & 36.24 & 10.00 \\
        DAViD
        & 528.9 & 571.8 & 553.7
        & 39.39 & 46.95 & 44.00 & 45.12
        & 37.00 & 10.00 \\
        Dex4D
        & 663.8 & 756.4 & 717.2
        & 40.60 & 47.01 & 46.49 & 46.54
        & 29.29 & 10.00 \\
        \midrule
        DeVI
        & \textbf{27.23} & \textbf{29.75} & \textbf{28.69}
        & \textbf{68.81} & \textbf{80.85}
        & \textbf{77.71} & \textbf{78.36}
        & \textbf{83.33} & \textbf{90.00} \\
        \bottomrule
    \end{tabular}
    \captionof{table}{\textbf{Quality of Video Imitation.} We evaluate the imitation quality on synthetic videos.}
    \label{tab:quant_video}
\end{minipage}\hfill
\begin{minipage}[t]{0.42\linewidth}
    \vspace{0pt}
    \centering
    \scriptsize
    \setlength{\tabcolsep}{0.8pt}
    \renewcommand{\arraystretch}{1.25}
    \begin{tabular}{@{}lcccccc@{}}
        \toprule
        \multirow{2}{*}{\raisebox{-2.0pt}{Methods}} &
        \multicolumn{2}{c}{Vertex Err. $\downarrow$} &
        \multicolumn{2}{c}{Rotation Err. $\downarrow$} &
        \multirow{2}{*}{\raisebox{-12.0pt}{\shortstack{Position Err.\\(m) $\downarrow$}}} \\
        \cmidrule(lr){2-3} \cmidrule(lr){4-5}
        & 2D (px) & 3D (m) & 2D (deg.) & 3D (deg.) & \\
        \midrule
        \shortstack[l]{RIGVid}
            & 235.6 & 0.4023 & 54.63 & 83.12 & 0.3955 \\
        \shortstack[l]{DAViD}
            & 10.40 & 0.2808 & 28.10 & \textbf{47.91} & 0.2763 \\
        \shortstack[l]{Dex4D}
            & 19.01 & \textbf{0.2277} & 30.67  & 59.10 & \textbf{0.2237} \\
        \midrule
        DeVI
            & \textbf{6.945} & N/A & \textbf{26.49} & N/A & N/A \\
        \bottomrule
    \end{tabular}
    \captionof{table}{\textbf{Representation Choice.} We evaluate the quality of object motion recovery.}
    \label{tab:representation_choice}
\end{minipage}
\vspace{-15pt}
\end{table}

%% file: contents/02_qual_video_imitation.tex
\begin{figure}[t]\centering
\includegraphics[width=\linewidth, trim={0 0 0 0},clip]{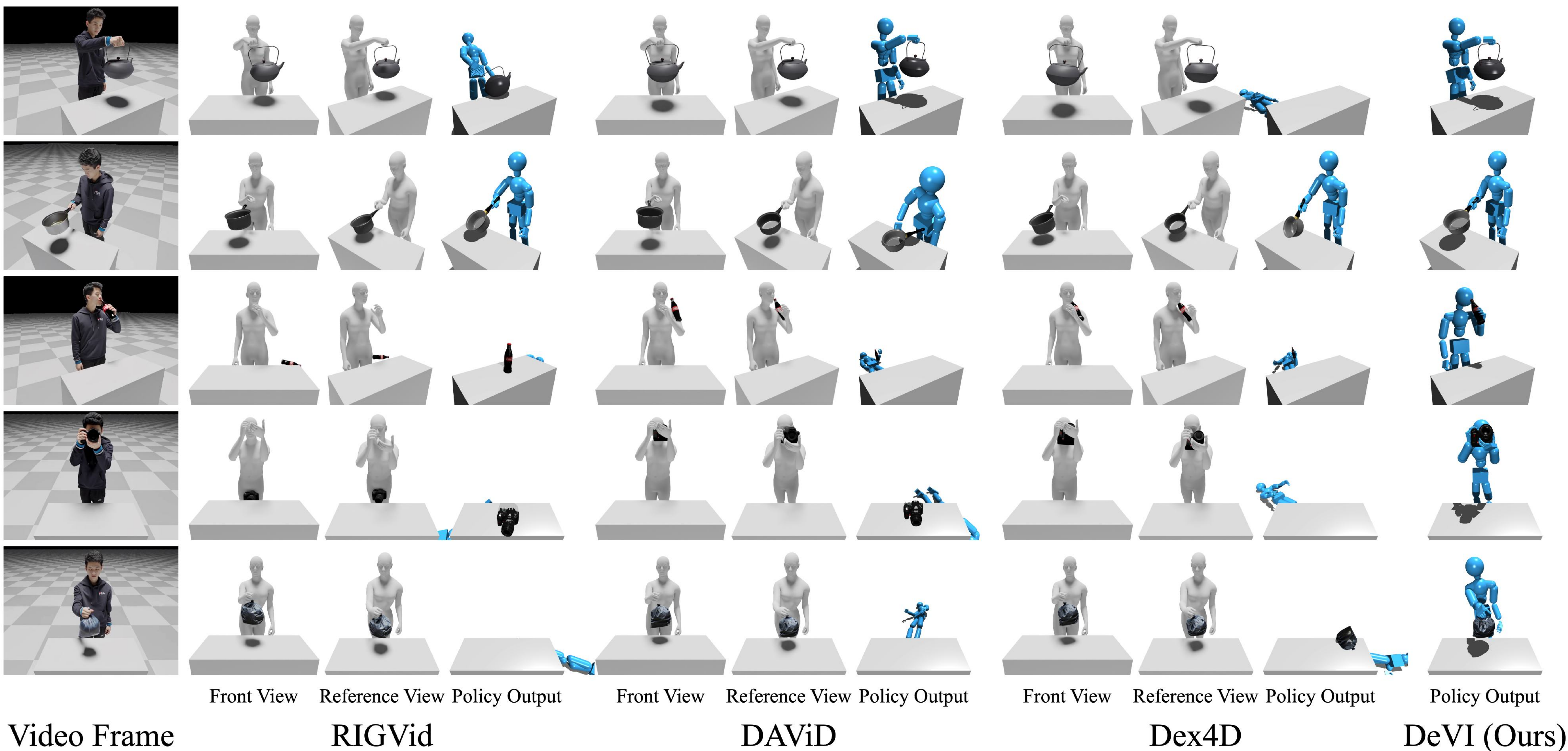}
\vspace{-15pt}
\caption{\textbf{Qualitative Comparison with Baselines.} DeVI avoids reconstructing unreliable 3D object trajectories compared to baselines which show significant human-object misalignments.
}
\label{fig:qual2}
\vspace{-8pt}
\end{figure}

%% file: contents/04_quant_mocap.tex
\begin{table}[!t]
\centering
\begin{minipage}[t]{0.60\linewidth}
    \vspace{0pt}
    \centering
    \scriptsize
    \setlength{\tabcolsep}{6.7pt}
    \renewcommand{\arraystretch}{1.02}
    \begin{tabular}{@{}lcccccc@{}}
        \toprule
        \multirow{2}{*}{\raisebox{-2.0pt}{Methods}} &
        \multicolumn{3}{c}{MPJPE (mm) $\downarrow$} &
        \multirow{2}{*}{\raisebox{-12.0pt}{\shortstack{$T_{\text{root}}$\\(mm) $\downarrow$}}} &
        \multirow{2}{*}{\raisebox{-13.0pt}{\shortstack{$T_{\text{obj}}$\\(mm) $\downarrow$}}} &
        \multirow{2}{*}{\raisebox{-2.0pt}{$O_{\text{obj}} \downarrow$}} \\
        \cmidrule(lr){2-4}
        & Body & Hand & All & & & \\
        \midrule
        PhysHOI & 93.62 & 120.9 & 109.5 & 85.09 & 79.80 & 0.9157 \\
        DeVI & \textbf{27.78} & \textbf{21.18} & \textbf{23.96} & \textbf{29.54} & \textbf{21.04} & \textbf{0.8855} \\
        \midrule
        SkillMimic & 108.2 & 150.4 & 134.4 & 100.4 & 83.80 & 1.130 \\
        DeVI & \textbf{34.08} & \textbf{29.56} & \textbf{31.46} & \textbf{37.62} & \textbf{24.88} & \textbf{1.023} \\
        \midrule
        InterMimic & 58.58 & 105.8 & 87.63 & 47.80 & 90.82 & 1.207 \\
        DeVI & \textbf{37.29} & \textbf{37.67} & \textbf{37.52} & \textbf{38.05} & \textbf{26.45} & \textbf{0.9173} \\
        \midrule
        $\text{DeVI}_{\text{w/o 2D Reward}}$ & 70.76 & 90.16 & 81.97 & 68.12 & 70.35 & 0.8195 \\
        DeVI & \textbf{37.53} & \textbf{31.80} & \textbf{34.21} & \textbf{41.67} & \textbf{20.89} & \textbf{0.6022} \\
        \bottomrule
    \end{tabular}
    \vspace{-5pt}
    \captionof{table}{\textbf{Quality of MoCap Imitation.} We evaluate the imitation accuracy of sequences in GRAB dataset.}
    \label{tab:quant_mocap}
\end{minipage}\hfill
\begin{minipage}[t]{0.37\linewidth}
    \vspace{0pt}
    \centering
    \scriptsize
    \setlength{\tabcolsep}{20.0pt}
    \renewcommand{\arraystretch}{1.15}
    \begin{tabular}{@{}l@{\hspace{1pt}}c@{\hspace{1pt}}c@{}}
        \toprule
        \multirow{2}{*}{\raisebox{-2.0pt}{Methods}} &
        \multicolumn{2}{c}{Success Ratio (\%) $\uparrow$} \\
        \cmidrule(lr){2-3}
        & $O_{\text{obj}}<1.0$ & $O_{\text{obj}}<2.0$ \\
        \midrule
        PhysHOI & 6.7 & 10.0 \\
        SkillMimic & 3.3 & 3.3 \\
        InterMimic & 16.7 & 43.3 \\
        $\text{DeVI}_{\text{w/o 2D Reward}}$ & 20.0 & 20.0 \\
        DeVI & \textbf{46.7} & \textbf{63.3} \\
        \bottomrule
    \end{tabular}
    \vspace{-5pt}
    \captionof{table}{\textbf{MoCap Success Ratio.} We evaluate the success ratio of the imitation with two different thresholds of rotation metric $O_{\text{obj}}$ along with $\text{MPJPE (All)} < 0.1$ and $T_{\text{obj}} < 0.1$.}
    \label{tab:quant_mocap_success}
\end{minipage}
\vspace{-15pt}
\end{table}

%% file: contents/06_visual_hoi_ablation_application.tex
\begin{figure}[!t]
\centering
\begin{minipage}[c]{0.60\linewidth}
    \centering
    \scriptsize
    \setlength{\tabcolsep}{2.5pt}
    \renewcommand{\arraystretch}{1.29}
    \begin{tabular}{@{}lccccccccc@{}}
        \toprule
        \multirow{2}{*}{\raisebox{-2.0pt}{Methods}} &
        \multicolumn{3}{c}{MPJPE (px) $\downarrow$} &
        \multirow{2}{*}{\raisebox{-12.0pt}{\shortstack{IoU\\(\%) $\uparrow$}}} &
        \multirow{2}{*}{\raisebox{-12.0pt}{\shortstack{Precision\\(\%) $\uparrow$}}} &
        \multirow{2}{*}{\raisebox{-12.0pt}{\shortstack{Recall\\(\%) $\uparrow$}}} &
        \multirow{2}{*}{\raisebox{-12.0pt}{\shortstack{F1\\(\%) $\uparrow$}}} &
        \multicolumn{2}{c}{Success (\%) $\uparrow$} \\
        \cmidrule(lr){2-4}
        \cmidrule(lr){9-10}
        & Body & Hand & All & & & & & Frame & Clip \\
        \midrule
        $\text{DeVI}_{\text{w/o VHA}}$
        & 94.04 & 90.19 & 91.82
        & 39.76 & 48.65 & 45.62 & 46.62
        & 40.57 & 10.00 \\
        DeVI
        & \textbf{34.21} & \textbf{11.86} & \textbf{21.32}
        & \textbf{68.81} & \textbf{80.85} & \textbf{77.71} & \textbf{78.36}
        & \textbf{83.33} & \textbf{90.00} \\
        \bottomrule
    \end{tabular}
    \captionof{table}{\textbf{Quantitative Ablation on Visual HOI Alignment.} We compare the policy learning results of DeVI with and without Visual HOI Alignment. 
    The correction of hand-object misalignment significantly improves
    both the motion quality and imitation success rates of the final policy.
    }
    \label{tab:visual_hoi_ablation}
\end{minipage}\hfill
\begin{minipage}[c]{0.36\linewidth}
    \centering
    \includegraphics[width=\linewidth]{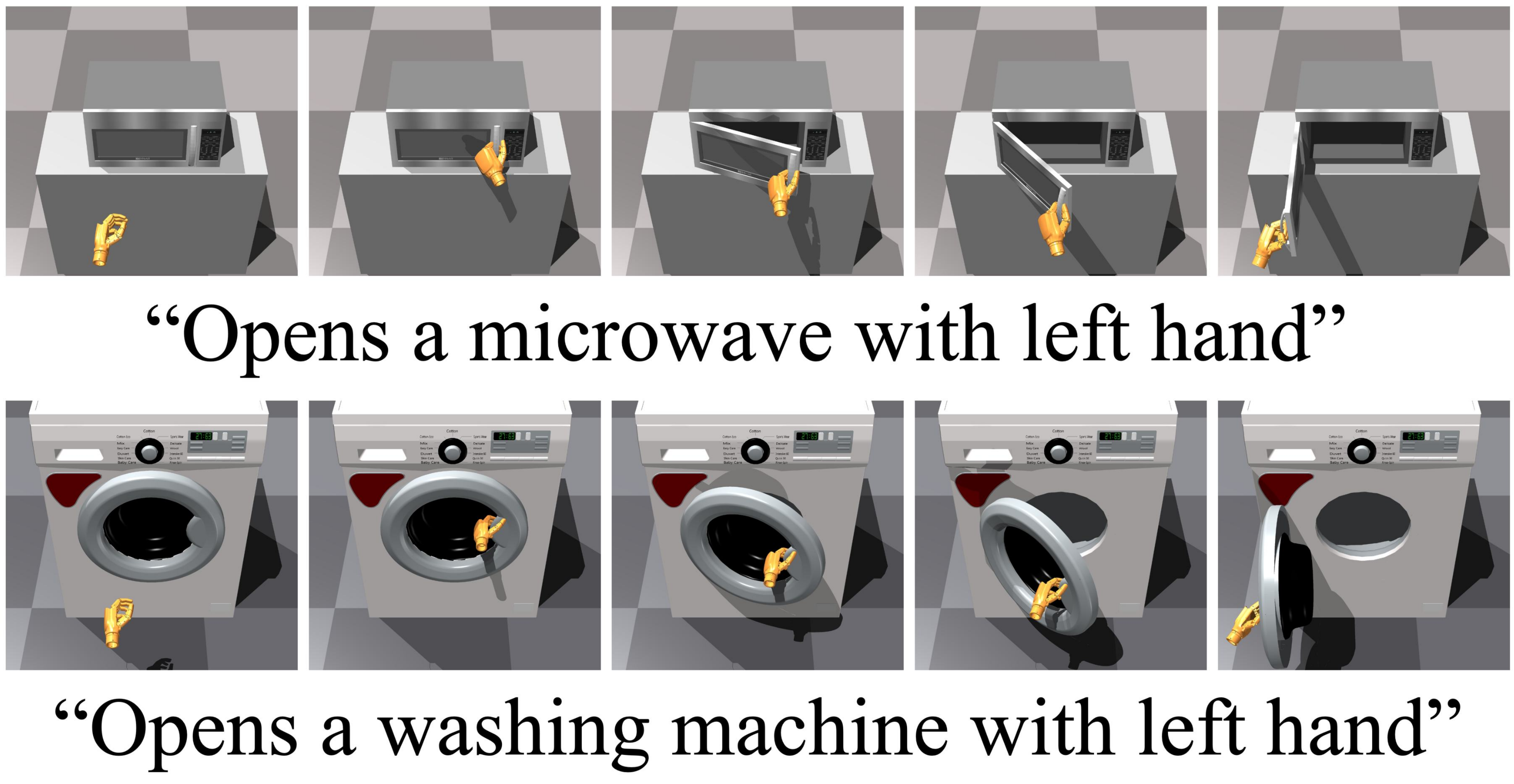}
    \vspace{-17pt}
    \captionof{figure}{\textbf{Articulated Object Manipulation.} DeVI can be extended to articulated object manipulation using egocentric video generation.
    }
    \label{fig:application}
\end{minipage}
\vspace{-7pt}
\end{figure}

%% file: contents/05_visual_hoi.tex
\begin{figure}[!t]
\centering
\captionsetup{justification=raggedright,singlelinecheck=false}
\begin{minipage}[t]{0.60\linewidth}
    \vspace{0pt}
    \centering
    \includegraphics[width=\linewidth]{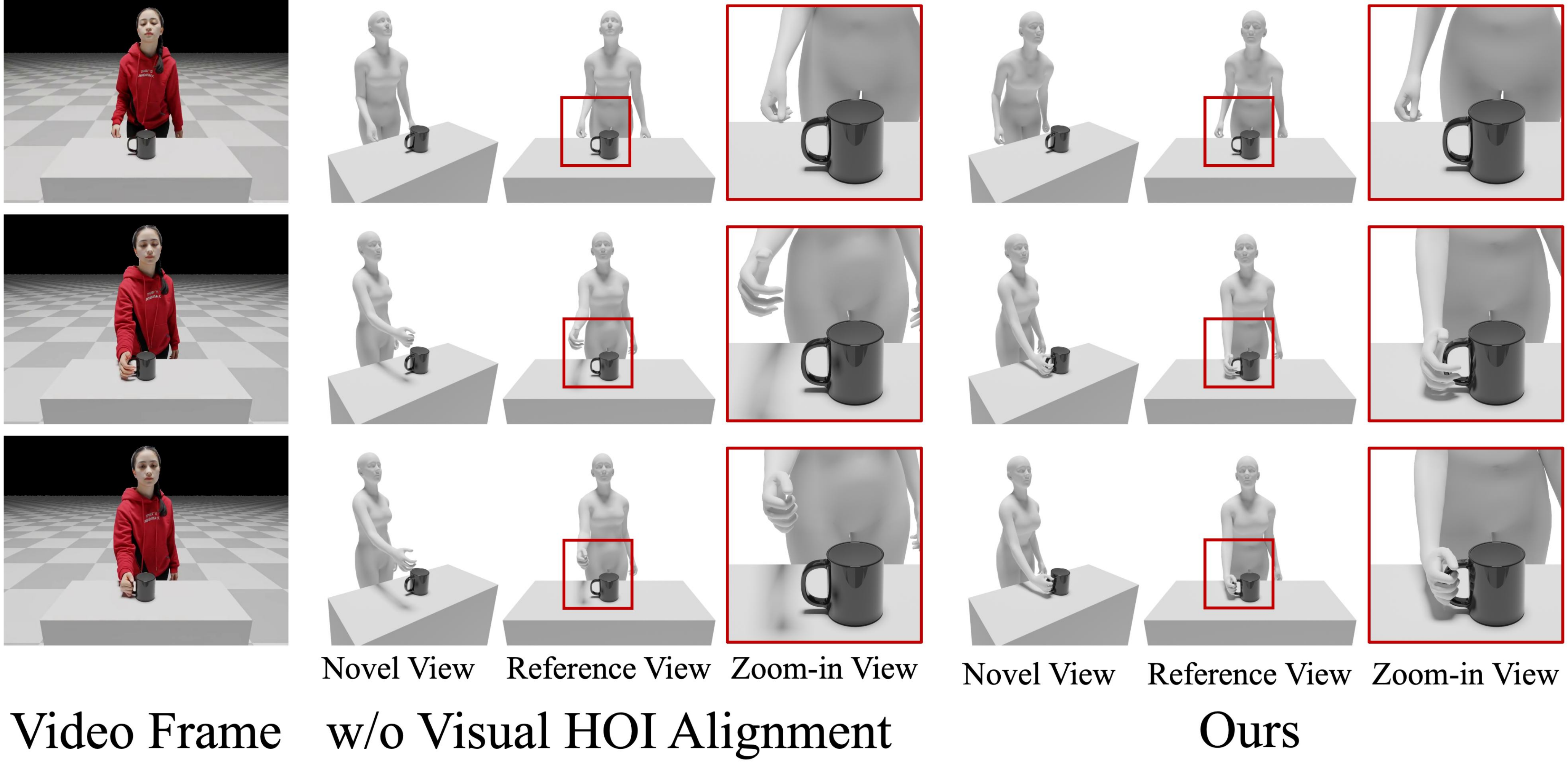}
    \vspace{-14pt}
    \captionof{figure}{\textbf{Qualitative Ablation on Visual HOI Alignment.}
    Visual HOI Alignment corrects reconstructed human motion to match the video and the initial 3D object configuration.
    }
    \label{fig:visual_hoi}
\end{minipage}\hfill
\begin{minipage}[t]{0.36\linewidth}
    \vspace{0pt}
    \centering
    \scriptsize
    \setlength{\tabcolsep}{5.2pt}
    \renewcommand{\arraystretch}{1.088}
    \begin{tabular}{@{}lccc@{}}
        \toprule
        \multirow{2}{*}{\raisebox{-2.0pt}{Methods}} & \multicolumn{3}{c}{MPJPE (mm) $\downarrow$} \\
        \cmidrule(lr){2-4}
        & Body & Hand & All \\
        \midrule
        TRAM & 387.9 & -- & -- \\
        PromptHMR & 75.2 & 94.2 & 85.9 \\
        PromptHMR + HaMeR & 75.2 & 101.4 & 90.3 \\
        GVHMR & 55.7 & 88.6 & 75.2 \\
        GVHMR + HaMeR & 55.7 & 89.9 & 76.0 \\
        \midrule
        DeVI & \textbf{53.9} & \textbf{69.4} & \textbf{63.1} \\
        \bottomrule
    \end{tabular}
    \captionof{table}{\textbf{Comparison with HMR.} We evaluate the quality of human mesh recovery especially in HOI manipulation scenario, showing the efficacy of Visual HOI Alignment.}
    \label{tab:hmr_arctic}
\end{minipage}
\vspace{-15pt}
\end{figure}

%% file: sections/06_discussion.tex
\section{Discussion}
\label{sec:discussion}

We present DeVI, which learns physics-based character control for dexterous HOI from text-conditioned synthetic videos without requiring 3D HOI MoCap demonstrations.
Inaccurate object motion estimates and hand-object misalignment make reliable imitation targets difficult to obtain.
To address this, our hybrid imitation targets bypass explicit 3D object motion reconstruction, allowing policy learning to find physically feasible object trajectories consistent with the 2D tracks.
Visual HOI Alignment further refines the human reference to match the video and initial 3D object configuration, improving reconstruction quality and imitation success.
Experiments demonstrate the benefits of hybrid tracking in both video and MoCap imitation.
Its effectiveness even with accurate 3D references suggests advantages beyond avoiding reconstruction noise.
Our functional manipulation results and egocentric applications demonstrate how generated videos can serve as motion plans for dexterous physical interaction in simulation and the real world.

%% file: sections/99_appendix.tex
\raggedbottom

\section{Implementation Details}
\label{sec:implementation_detail}

\subsection{Scene Initialization}
\label{subsec:scene_init}
Our scene initialization follows a tabletop scenario. 
We place an SMPL-X~\citep{SMPLX} human at the origin on the $xy$-plane and initialize it to face the positive y-axis.
We then place a table on the $xy$-plane at $(x, y) = (0.0, 0.4)$, with dimensions $(1.0, 0.5, 0.8)$.
The objects are initialized on the table to construct our scene. 
We assume a physically valid state (\eg, no floating objects) for the initialized scene and assign an initial pose to each object accordingly, using the same poses to initialize the physics simulation.

\subsection{Deforming Textured Human}
\label{subsec:rigging}
We replace the SMPL-X model with a 3D textured human mesh by deforming the mesh to match the pose of the initialized SMPL-X.
We use the 3D textured human meshes from the THuman 2.0 dataset~\citep{THuman2.0} with their corresponding SMPL-X annotations to assign SMPL-X offsets and skinning weights to each vertex of the textured human mesh. For the textured human mesh vertices $\{ x_i \}_{i=1}^N \in \mathbb{R}^{N\times3}$ and $\{ v_j \}_{j=1}^{M} \in \mathbb{R}^{10475\times3}$, we assign the offset $o_i \in \mathbb{R}^3$ and skinning weight $w_i \in \mathbb{R}^{J}$ to the vertex $x_i$ as follows:
\begin{align*}
    w_i = \sum_{k \in \mathcal{N}_K(i) }\alpha_{i,k} W_{k}, \quad o_i &= \sum_{k \in \mathcal{N}_K(i) }\alpha_{i,k} O_{k}, 
\end{align*}
where $\mathcal{N}_K(i)$ is the set of indices of the K-nearest vertices of $x_i$ in $\{ v_j \}_{j=1}^{M}$, $W_k \in \mathbb{R}^{J}$ is the skinning-weight vector for the $J=55$ joints at SMPL-X vertex $v_k$, $O_k \in \mathbb{R}^3$ is the offset of the SMPL-X vertex $v_k$ computed via shape and pose parameters, and $\alpha_{i,k}$ is the blending coefficient of each skinning weight.
The coefficient $\alpha_{i,k}$ is defined using a Gaussian kernel over the distance between $x_i$ and its nearest neighbors $\{ v_k \}_{k \in \mathcal{N}_K(i)}$ as follows:
\begin{align*}
    \alpha_{i,k} = \frac{ e^{s_{i, k}} }{\sum_{k' \in \mathcal{N}_K(i)} e^{s_{i, k'}}}, \quad
    s_{i, k}  = -\frac{\| x_i - v_k \|^2}{2\sigma^2}
\end{align*}
Using the assigned offset $o_i$ and skinning weight $w_i$, we deform the 3D textured human mesh via linear blend skinning.
In practice, we use $K=16$ for approximating the offsets and skinning weights.

\subsection{Rendering Scene}
For the initialized scene, we initially place 16 candidate cameras at a radius of 1.5, spaced at $45\degree$ azimuth intervals, and facing the origin with two elevation angle variations, $15\degree$ and $30\degree$.
The height of the cameras is adjusted about 1.0 along the positive z-axis to capture the scene within the image frame.
Among the 16 cameras, we empirically find that selecting one of the six frontal views in which both the human’s hands and the object are visible is beneficial, especially for hand pose estimation.
For the remaining intrinsic parameters, we follow Isaac Gym~\citep{isaacgym}, and define the intrinsic matrix $K$ as follows:
\begin{equation}
    K = \begin{bmatrix}
        f & 0 & w/2 \\
        0 & f & h/2 \\
        0 & 0 & 1 
    \end{bmatrix},
\end{equation}
where $f = w / 2$ and $h$, $w$ are the height and width of the image. In practice, we use $h=576$, $w=1024$ for rendering.
The installed camera is used to render an image, which is used as the input for generating motion plans in the form of a video using a video diffusion model.

\subsection{Generating HOI Video}
To generate an HOI video from the rendered images, we use a pre-trained video diffusion model, Wan~\citep{Wan} with additional LightX2V~\citep{lightx2v} LoRA for faster inference.
Generating a single video takes about 10 minutes on an NVIDIA A6000 GPU.
For the text prompt, we use the following format:
\begin{tcolorbox}[colback=gray!5, colframe=gray!50, fonttitle=\bfseries]
\ttfamily\small
A person [action] [object] with [hand index] [detail]. Both hands are visible in the frame. Stationary camera.
\end{tcolorbox}
We use both hand-designed prompts and prompts automatically generated by ChatGPT~\citep{ChatGPT} to fill in the brackets.
While keeping the format, we additionally adjust the word ``person'' and gendered pronouns in the prompt according to the gender of the 3D textured human mesh, and add details about the non-interacting hand.

\subsection{Unifying Body and Hand Estimators}
\label{subsec:unify_body_hand}
To reconstruct the coarse human from the generated video, we use the body estimator GVHMR~\citep{gvhmr} and the hand estimator HaMeR~\citep{HaMeR}.
To unify the outputs of each estimator into a single SMPL-X, we replace the body estimator’s local wrist pose with the following transformation:
\begin{align*}
    G_e &=\prod_{r=0}^{e}R_{p_r} \\
    R_w &= G_e^{-1}G_w
\end{align*},
where $R_{p_r} \in SO(3)$ is the rotation of each body joint $p_r$ obtained from the body estimator, and $G_w$ is the global rotation of the hand (in this case, the wrist) obtained from the hand estimator. We define $(p_0,p_1,\dots,p_e)$ as an ordered sequence of joints along the forward-kinematics chain from the pelvis to the elbow. We omit the left/right subscripts for simplicity. The local rotation of the wrist $R_w$ is converted to axis-angle representation and used for the unified SMPL-X's wrist pose. 

\begin{figure*}[t]\centering
\includegraphics[width=\linewidth, trim={0 0 0 0},clip]{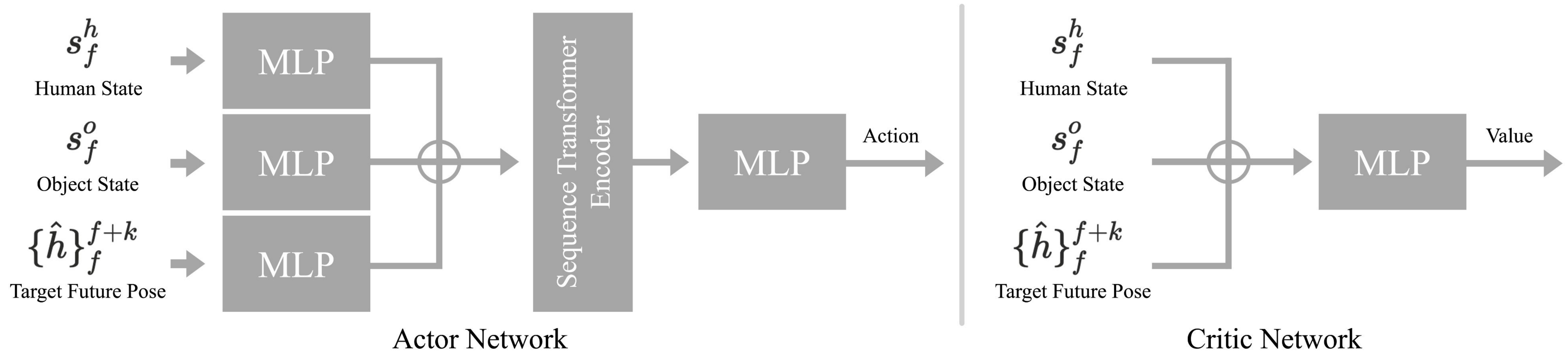}
\captionof{figure}{\textbf{Network Architecture of DeVI.} Our humanoid control policy network consists of a transformer-based actor network and an MLP-based critic network with the same input states.
}
\label{fig:network}
\end{figure*}

\subsection{Details of Visual HOI Alignment}
\label{subsec:details_visual_hoi}
To reconstruct human motion aligned with both the video and the existing 3D object, our Visual HOI alignment optimizes SMPL-X body and hand parameters using a 2D projection loss and a one-sided Chamfer distance loss.
Even though we compute the loss using 19 body joints and 32 hand joints, optimizing all body parameters adversely affects the estimates for occluded joints (especially in the lower body).
In practice, we optimize only the upper-body poses, specifically for the hands, wrists, elbows, shoulders, and spine joints. 
We use the Adam~\citep{adam} optimizer with a learning rate of $2 \times 10^{-2}$ without decay.
For each loss, we use $w_b = 1.0$,  $w_h = 1.0$,  $w_{tc} = 1.0\times 10^{4}$, and $w_{\text{HOI}} = 5.0\times 10^{2}$.

\subsection{Contact Estimation}
\label{subsec:contact_estimation}
Following GRAIL~\citep{GRAIL}, we annotate human-object contact from video frames using a vision-language model (VLM).
For each frame, we provide the VLM~\citep{gpt6-astra} with the image and the target object name to obtain a contact decision and confidence level in JSON format.
The prompt defines contact as physical touching by any body part, excludes proximity without touching, and instructs the model to return false when uncertain.
We set the binary pseudo contact label $\psi_t$ to $1$ only when \texttt{in\_contact} is \texttt{true} and \texttt{confidence} is \texttt{high}, and to $0$ otherwise.
These global labels provide contact timing cues, while the contact reward is evaluated only for participating hands, as detailed in Appendix~\ref{subsec:contact_reward_details}.
For every frame, we use the following prompt, replacing \texttt{\{object\_name\}} with the target object name.
\begin{tcolorbox}[colback=gray!5, colframe=gray!50]
\ttfamily\small\raggedright
You are a careful vision classifier. Given an image of a person and an object, determine whether the person is in physical contact with the object.\par
Object: \{object\_name\}\par
Guidelines:\par
- Contact means the person is physically touching or holding the object.\par
- Near proximity without touching does NOT count as contact.\par
- If any part of the person's body touches the object, return true.\par
- Be conservative: when uncertain, return false.\par
Respond with strictly valid JSON using this schema (no extra text):\par
\{"in\_contact": true/false, "confidence": "high/medium/low"\}
\end{tcolorbox}

\subsection{Network Architecture}
\label{subsec:architecture}
Our humanoid control policy network uses an actor-critic architecture.
Both the actor and critic networks take the human state, object state, and target future pose as inputs.
Each input of the actor network first passes through a separate 2-layer MLP with 256 hidden units.
The resulting features are then concatenated and fed into a sequence transformer encoder for encoding.
The encoded latent passes through a 3-layer MLP with 1024 hidden units and outputs the action of the humanoid.
In contrast, the three inputs of the critic network are flattened and concatenated without a separate encoding stage, and pass through a 4-layer MLP with 1024 hidden units to output the value of the given state. 
The overall architecture is shown in Fig.~\ref{fig:network}.

\subsection{Details about Hybrid Tracking Reward}
\label{subsec:hybrid_reward_details}

Each of our tracking term shares the same exponential form
\begin{equation}
r_k \;=\; \exp\!\big(c_k \cdot \bar e_k\big),
\end{equation}
where $\bar e_k$ is the tracking error between the simulated state and the reference $\hat h$ (a mean squared distance for position terms and a mean geodesic angular distance for rotation terms), and $c_k<0$ is a sensitivity coefficient.
The error terms and coefficients used in our experiments are summarized in Table~\ref{tab:reward_term}. For the full-body terms ($r_{\text{jp}}, r_{\text{jv}}, r_{\text{jr}}$), positions, velocities, and rotations are expressed in the global frame and averaged over all body links. For the local hand terms ($r_{\text{lp}}^{h}, r_{\text{lr}}^{h}$), finger keypoints are first translated to the wrist origin and rotated into the wrist frame. 
The power penalty $r_{\text{pw}}$ follows~\citep{QuestEnvSim} and discourages excessive joint torques, with $\tau_j$ and $\dot q_j$ denoting the actuator torque and velocity at the $j$-th joint.
In addition, we redundantly apply the same exponential form to the fingertip locations ($T\subset H$) and to a small subset of key body links ($K$, the two wrists and two ankles) on top of $r_{\text{lp}}^{h}$ and $r_{\text{jp}}$, with coefficients $-100$ and $-120$ respectively, to place stronger emphasis on tracking these contact-relevant points.
For the 2D object tracking reward, we use $\lambda_o = -1.0 \times 10^{-3}$ as a coefficient.

For the DeVI variant without the 2D reward in Tabs.~\ref{tab:quant_mocap} and~\ref{tab:quant_mocap_success}, we replace $R_o$ with the following 6D object pose tracking reward.
\begin{equation}
    R_{\text{6D}} = \exp\!\left(c_{\text{pos}}\lVert \vec{p}_t - \vec{p}_t^{\text{ref}}\rVert_2^2\right)\exp\!\left(c_{\text{rot}}\theta_t^2\right),
    \label{eq:6d_object_reward}
\end{equation}
where $\vec{p}_t$ and $\vec{p}_t^{\text{ref}}$ denote the simulated and reference object positions in meters, respectively, and $\theta_t$ is the angle of their relative rotation in radians.
We use $c_{\text{pos}}=-80$ and $c_{\text{rot}}=-1$ in all 6D object tracking comparisons.
For the generated-video comparisons, we use the same reward and coefficients with reconstructed object trajectories from RIGVid~\citep{RigVid}, DAViD~\citep{DAViD}, and Dex4D~\citep{Dex4D}.
In the DeVI ablation, the human reference, other rewards, network architecture, training budget, initialization, and human termination criteria remain identical.
Only the object target, tracking reward, and object termination criterion change, as detailed in Appendix~\ref{subsec:training_details}.

\begin{table}[t]
    \small
    \centering
    \vspace{-5pt}
    \renewcommand{\arraystretch}{1.4}
    \resizebox{0.6\linewidth}{!}{
        \begin{tabular}{llc}
        \toprule
        Term & Error $\bar e_k$ & Coefficient $c_k$ \\
        \midrule
        $r_{\text{jp}}$ (body joint position) & $\tfrac{1}{B}\sum_b \lVert p_b - \hat p_b\rVert_2^{2}$ & $-100$ \\
        $r_{\text{jv}}$ (body joint velocity) & $\tfrac{1}{B}\sum_b \tfrac{1}{3}\lVert v_b - \hat v_b\rVert_2^{2}$ & $-0.5$ \\
        $r_{\text{jr}}$ (body joint rotation) & $\tfrac{1}{B}\sum_b \theta(q_b,\hat q_b)$ & $-1.5$ \\
        $r_{\text{lp}}^{h}$ (local hand position) & $\tfrac{1}{H}\sum_h \tfrac{1}{3}\lVert p_h^{w} - \hat p_h^{w}\rVert_2^{2}$ & $-100$ \\
        $r_{\text{lr}}^{h}$ (local hand rotation) & $\tfrac{1}{H}\sum_h \theta(q_h^{w},\hat q_h^{w})$ & $-2.0$ \\
        $r_{\text{pw}}$ (power penalty) & $\sum_{j=1}^{D} |\tau_j\,\dot q_j|$ & $-2.0\!\times\!10^{-5}$ \\
        \bottomrule
        \end{tabular}
    }
    \caption{\textbf{Reward Term Definitions.} Error definitions and coefficients for each tracking term.}
    \label{tab:reward_term}
\end{table}

\subsection{Details about Contact Reward}
\label{subsec:contact_reward_details}
The contact reward, $R_{\text{contact}} = R_{cf} \cdot R_{cd}$, encourages the humanoid to establish contact with the target object in the simulation. It is defined via the product of the contact force reward $R_{cf}$ and the contact distance reward $R_{cd}$.
While similar formulations appear in previous studies~\citep{PhysHOI, InterMimic}, our reward relies on contact timing cues explicitly inferred from the generated video $\mathcal{V}$.
For each frame, we use the global binary contact annotation $\psi_t \in \{0,1\}$ from the VLM~\citep{gpt6-astra}, as described in Appendix~\ref{subsec:contact_estimation}.
For both contact terms, we include only the participating hands, denoted by $\mathcal{A} \subseteq \{L,R\}$, excluding the non-participating hand in single-hand manipulation.
Using this shared contact label, we average the per-hand contact force rewards as follows:
\begin{align}
    R_{\text{cf}} &= (1 - \psi_t) + \frac{\psi_t}{|\mathcal{A}|}\sum_{h\in\mathcal{A}}\Psi_t^h,
\end{align}
where $\Psi_t^h \in [0,1]$ is the ratio of force sensors in hand $h$ for which the measured force exceeds a predefined threshold as follows:
\begin{equation}
    \Psi_t^h = \frac{1}{K_h}\sum_{j=1}^{K_h} \mathbb{I}[ c_t^{h,j} > \tau_{\text{contact}} ],
\end{equation}
where $K_h$ is the number of force sensors in hand $h$, $c_t^{h,j} \in \mathbb{R}$ is the contact force measured at its $j$-th joint at simulation step $t$, $\tau_{\text{contact}}$ is the contact force threshold, and $\mathbb{I}$ is an indicator function.
Using the one-sided Chamfer distance $d_t^h$ from the joints of hand $h$ to object vertices, we average the per-hand contact distance rewards as follows:
\begin{align}
    R_{\text{cd}} &= (1 - \psi_t) + \frac{\psi_t}{|\mathcal{A}|}\sum_{h\in\mathcal{A}}\sigma\!\left(-\lambda_{\text{c}}(d_t^h)^2\right),
\end{align}
where $\sigma$ is a sigmoid function and $\lambda_{c}$ is a weighting factor.
In practice we use $\tau_{\text{contact}}=0.1$ and $\lambda_{c} = 10.0$ in our experiments.

\subsection{Training Details}
\label{subsec:training_details}
\noindent\textbf{Time Sampling for Initialization.} 
Previous studies on imitating human motions~\citep{MaskedMimic, SkillMimic} sample a random time step from the reference motion for initialization, which has a positive effect on sample efficiency during rollout.
Unlike these studies, we do not have access to the object’s 6D pose reference, which makes initialization at an arbitrary time step difficult.
Instead, we propose a strategy that initializes at the pre-contact frame where the object’s 6D pose is the same as the initial pose, with a 50\% probability.
This strategy comes from the observation that, in general, more trials are required near the frame where the human starts interacting with the object than in intervals that solely imitate human motion.
We empirically find that this strategy boosts policy learning compared to first-frame initialization.

\noindent\textbf{Early Termination.} 
For sample efficiency during rollout, we perform early termination when the current state is significantly different from the reference imitation target.
We define early termination based on the per-joint distance error for 3D body and fingertip joints, and the pixel distance error for the 2D object trajectory.
Specifically, we terminate if the mean error of the 3D body joints exceeds 200 mm or if any 3D body joint error exceeds 400 mm, or if the mean 3D fingertip error exceeds 40 mm.
For the pixel distance error of the 2D object trajectory, we define the threshold based on the image resolution as follows.
\begin{equation}
    \tau_{\text{2D}} = \alpha_{\text{2D}}\sqrt{W^2 + H^2}
\end{equation}
In practice, we use $\alpha_{2D}=0.08$, giving approximately $\tau_{2D}=94$ pixels for $W=1024$ and $H=576$.
For the 6D object tracking settings, we instead terminate when the mean distance between corresponding object points exceeds $0.15$~m.
\begin{equation}
    E_{\text{3D}}(t) = \frac{1}{N}\sum_{i=1}^{N}\left\lVert\vec{y}_{t,i}-\vec{y}_{t,i}^{\text{ref}}\right\rVert_2 > 0.15~\text{m}, \qquad N=1024,
\end{equation}
where $\vec{y}_{t,i}$ and $\vec{y}_{t,i}^{\text{ref}}$ are the same sampled object point transformed by the simulated and reference poses, respectively.
This error captures both translation and rotation and is used for termination, while the reward follows Eq.~\ref{eq:6d_object_reward}.

\noindent\textbf{Training Setup.} 
For training our actor-critic network, we use the Adam optimizer with a learning rate of $2 \times 10^{-5}$ for the actor network and $1 \times 10^{-4}$ for the critic network.
We collect data using 4096 environments in Isaac Gym, update each network’s parameters after 32 rollouts, and use a batch size of 1024.
For hardware, we train on NVIDIA A6000 GPUs, and it takes about 20 hours on a single GPU to imitate a 263-frame video, although the runtime varies by reference motion.

\noindent\textbf{Policy Learning.} 
To learn the humanoid control policy using our hybrid tracking reward, we build actor-critic network with transformer-based actor network and MLP-based critic network. The actor network outputs the action which is used as a control signal for the humanoid, and the critic network estimates the value function which is used to compute the advantage $A_t$ of the current state-action pair using GAE~\citep{GAE}. We update the actor network using the policy gradient from PPO~\citep{PPO} as follows: 
\begin{align}
    \mathcal{L}_{\text{ppo}} &= -\mathbb{E}_t [ \min ( r_t(\psi_{\text{actor}})A_t, \text{clip}( r_t(\theta), 1-\epsilon, 1+\epsilon ) A_t ] \\
    \mathcal{L}_{\text{bound}} &= \sum_i \left( \text{ReLU}(\mu_i - 1)^2 + \text{ReLU}(-\mu_i - 1)^2 \right),
\end{align}
where $\epsilon \in \mathbb{R}$ is a clipping constant, $\text{clip}(\cdot, a, b)$ is a clip function from $a$ to $b$, $r_t(\psi)$ is the ratio of the likelihoods of the current action under the updated and old policies, and $\mu_i$ is an output of the actor network which is the estimated mean of the distribution. We softly update the actor network so that the likelihood of actions with higher advantage increases, while ensuring that the output action remains bounded in $[-1, 1]$ via the following actor loss:
\begin{align}
    \mathcal{L}_{\text{actor}} = \mathcal{L}_{\text{ppo}} + \mathcal{L}_{\text{bound}}
\end{align}
Along with the actor loss, we update the critic network to predict the estimated returns, modelling a reliable value function as follows:
\begin{align}
    V_{\text{clip}} &= \text{clip}(V_{\psi_{\text{critic}}}, V_{\text{old}} - \epsilon, V_{\text{old}} + \epsilon) \\
    \mathcal{L}_{\text{critic}} &= \frac{1}{2}\mathbb{E} \left[ \max \left( \left( V_{\psi_{\text{critic}}} - R_t \right)^2, \left( V_{\text{clip}} - R_t \right)^2   \right) \right], 
\end{align}
where $V_{\text{old}}$ is an old value function, $V_{\psi_{\text{critic}}}$ is the current estimated value function, and $R_t = A_t + V_{\text{old}}$ is the target return. For the clipping constant, we use $\epsilon =0.2$.

\subsection{Penetration Issue in Rigid-Body Simulation}
\label{subsec:contact_resolution}
Penetration between the humanoid and the object in physics-based human-object interaction can arise from two distinct sources.
First, physics simulators commonly use simplified collision geometries that differ from the higher-resolution visual meshes used for rendering, so the simulator may correctly resolve contact between the underlying collision geometries while the visual meshes appear to overlap.
Second, rigid-body simulators~\citep{isaacgym, isaacsim, mujoco} resolve contacts via iterative constraint solvers under a fixed iteration budget, which inherently leaves contact constraints imperfectly satisfied.
The effect is amplified in dexterous manipulation, where many simultaneous finger-object contacts must be resolved within a single discrete timestep.
State-of-the-art physics-based HOI methods~\citep{PhysHOI, SkillMimic, InterMimic} also exhibit such residual penetration, as specifically noted by MaskedManipulator~\citep{MaskedManipulator}.
Following MaskedManipulator~\citep{MaskedManipulator}, we adopt a power penalty on the hand degrees of freedom as shown in Tab.~\ref{tab:reward_term}.

\section{Additional Results}
\label{sec:additional_results}

\subsection{Synthetic HOI Video Imitation}
\label{subsec:additional_video_imitation}

\begin{figure}[t]\centering
\includegraphics[width=\linewidth]{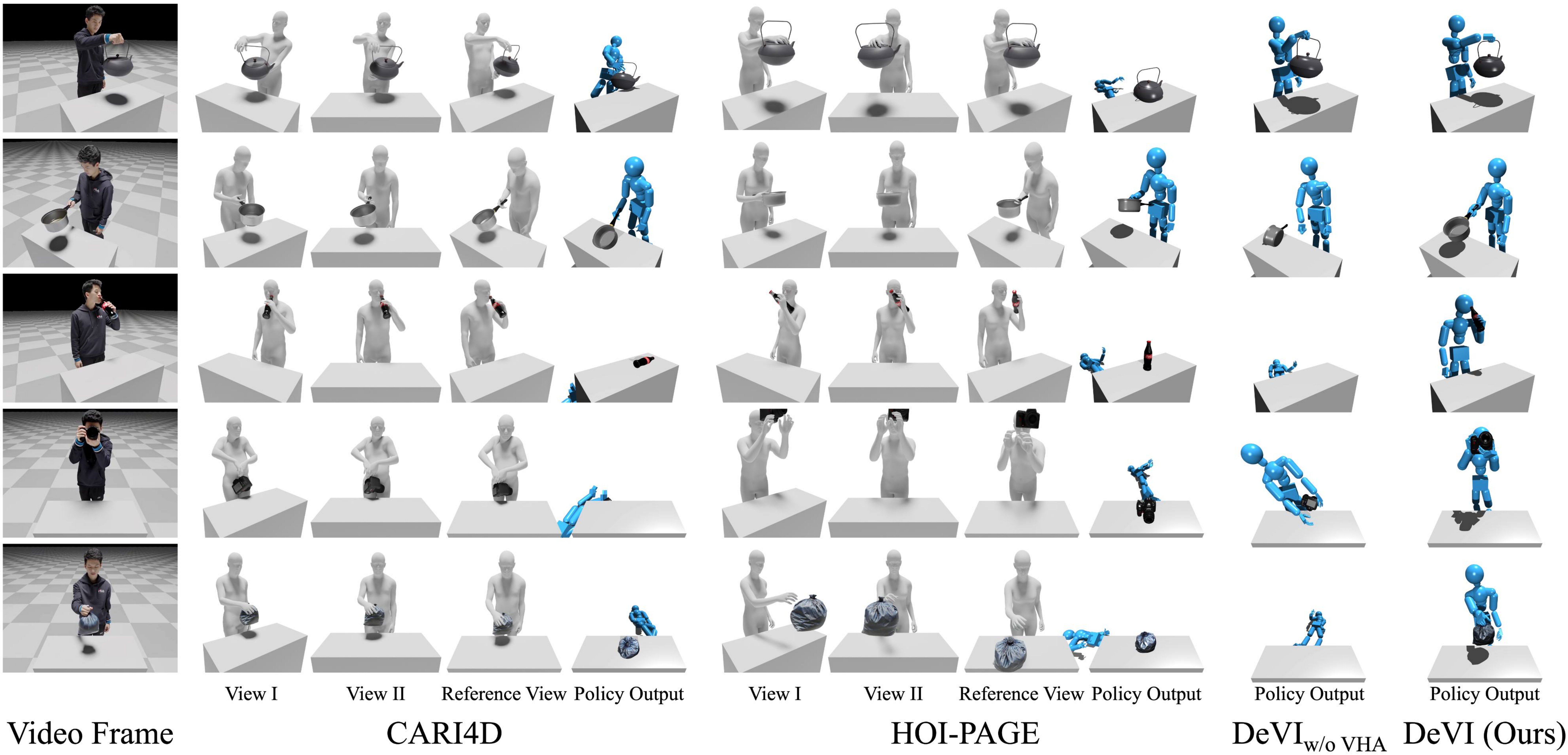}
\vspace{-15pt}
\caption{\textbf{Imitation from 4D HOI Reconstruction.} We compare CARI4D and HOI-PAGE with DeVI and the ablation without Visual HOI Alignment. Imprecise hand-object alignment leads to imitation failures, while DeVI successfully imitates the generated interactions.}
\label{fig:qual_4d_hoi}
\end{figure}

\noindent\textbf{Imitation from 4D HOI Reconstruction.}
To complement the object motion reconstruction comparison in Fig.~\ref{fig:qual2}, we evaluate CARI4D~\citep{CARI4D} and HOI-PAGE~\citep{HOIPAGE} using their reconstructed 4D human-object motions as imitation references.
As CARI4D and HOI-PAGE each reconstruct human motion from the video, the methods use different human references.
Following the evaluation in Tab.~\ref{tab:visual_hoi_ablation}, we therefore measure human MPJPE in image space using 2D joints predicted from the video as ground truth.
We also include DeVI without Visual HOI Alignment to evaluate the effect of correcting the human reference.
As shown in Fig.~\ref{fig:qual_4d_hoi}, both reconstruction methods fail to recover the precise hand-object configurations required for the shown dexterous interactions, leading to unsuccessful policy rollouts.
The ablation further shows that inaccurate human references hinder contact formation even when the object target remains in 2D.
As shown in Tab.~\ref{tab:quant_4d_hoi}, DeVI outperforms both reconstruction baselines across all metrics.

\input{contents/09_quant_4d_hoi}

\begin{figure}[t]\centering
\includegraphics[width=\linewidth]{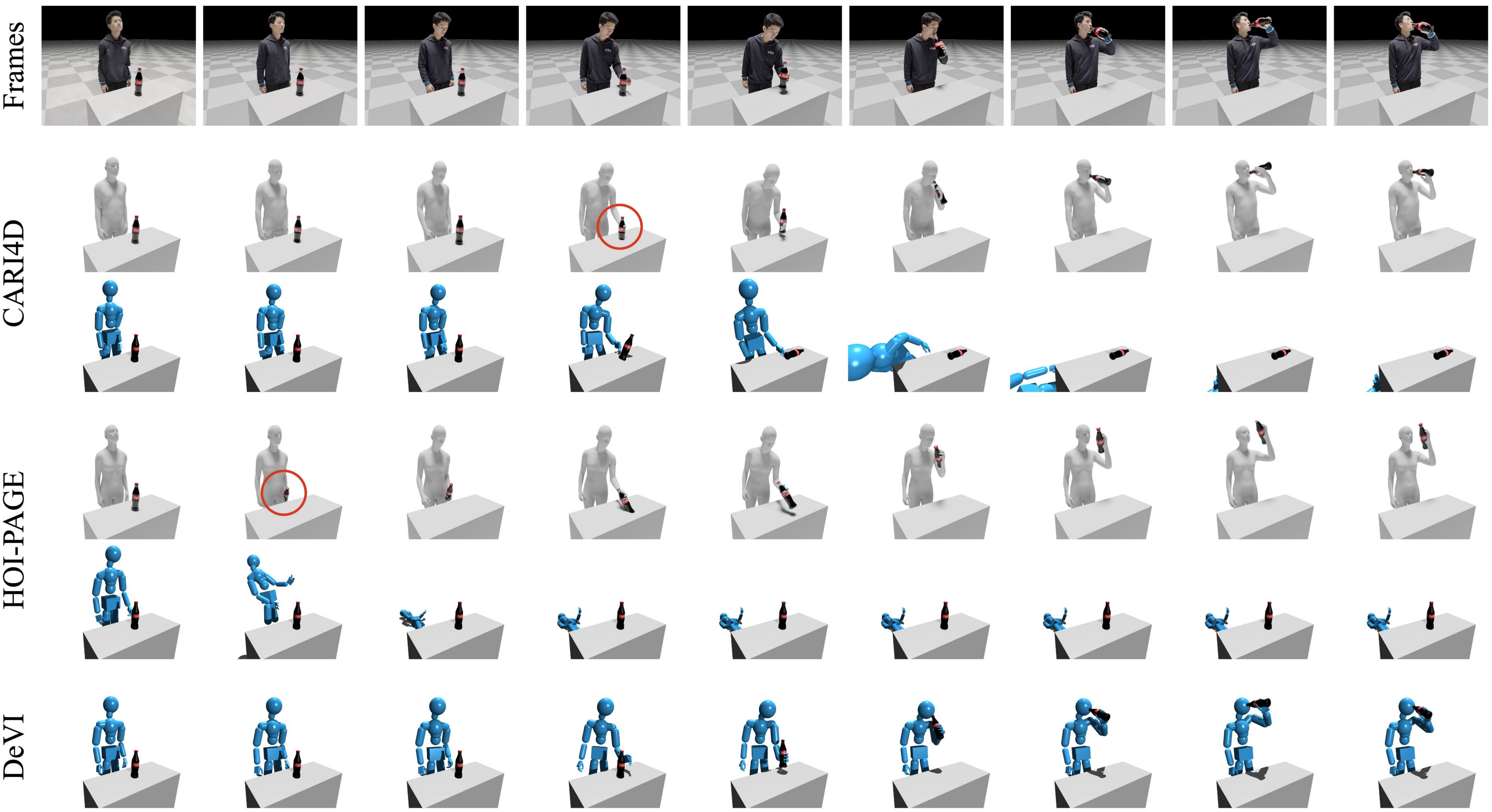}
\caption{\textbf{Failure Modes of 4D HOI Reconstruction Baselines.} HOI-PAGE produces noisy object positions, while CARI4D lacks the finger articulation needed for grasping despite more consistent human-object motion. Both references lead to imitation failures.}
\label{fig:qual_4d_hoi_failure}
\end{figure}

\noindent\textbf{Failure Analysis of 4D HOI Reconstruction.}
As shown in Fig.~\ref{fig:qual_4d_hoi_failure}, HOI-PAGE produces relatively noisy object position estimates over time, disrupting alignment between the object and the hand.
In the drinking sequence, the reconstructed bottle shifts before the hand establishes a grasp, providing an inconsistent reference for physical interaction.
While CARI4D recovers the spatio-temporal relationship between the human and object more consistently, the reconstructed finger articulation does not form the contacts required to grasp the object.
Consequently, the simulated hand reaches the bottle but fails to secure a grasp, causing the bottle to topple during imitation.
Using Visual HOI Alignment and hybrid imitation targets, DeVI establishes the required contacts and effectively imitates the generated interactions in the physics simulator without explicitly reconstructing the full 4D HOI motion.

\begin{figure*}[t]\centering
\includegraphics[width=\linewidth]{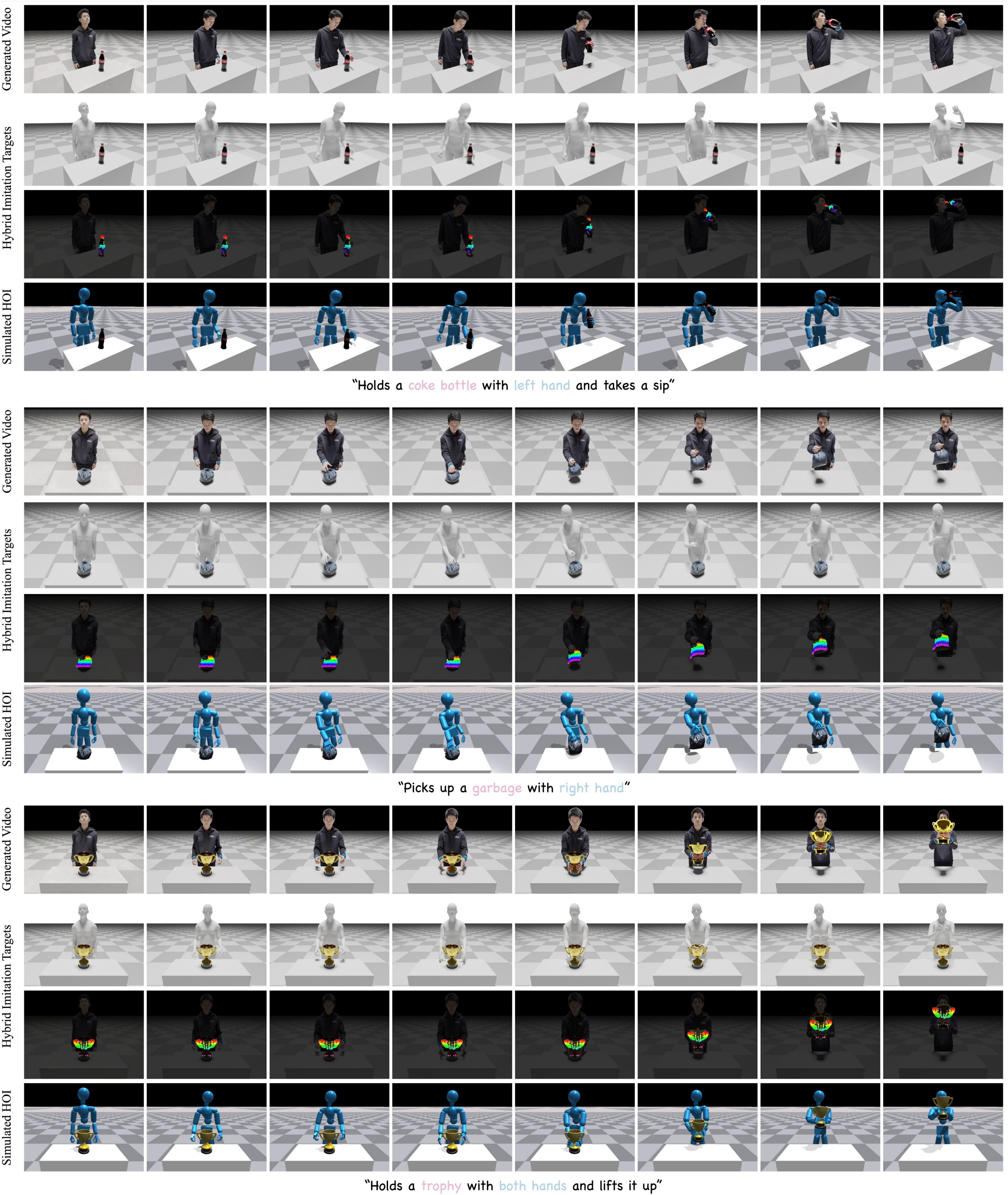}
\captionof{figure}{\textbf{Video-Motion Alignment.} 
}
\label{fig:P1}
\end{figure*}

\begin{figure*}[t]\centering
\includegraphics[width=\linewidth]{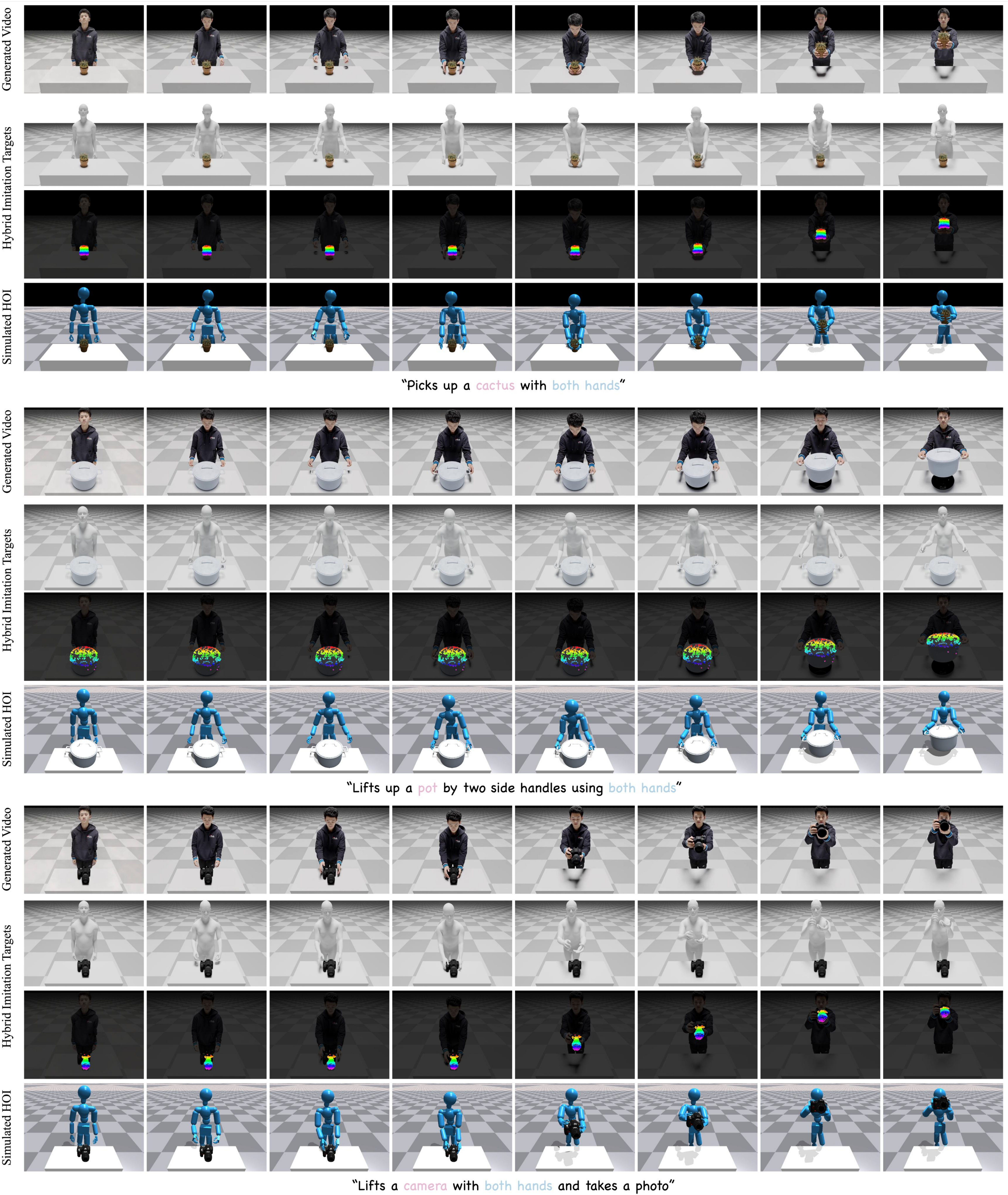}
\captionof{figure}{\textbf{Video-Motion Alignment.} 
}
\label{fig:P2}
\end{figure*}

\noindent\textbf{Detailed Results with Hybrid Imitation Targets.} Detailed results of the simulated HOIs, including the text prompts and intermediate hybrid imitation targets, are shown in Fig.~\ref{fig:P1} and Fig.~\ref{fig:P2}.
As shown in Fig.~\ref{fig:P1} and Fig.~\ref{fig:P2}, the simulated HOI motions produced by policies trained to imitate the hybrid imitation targets are well aligned with the generated videos.
We show that hybrid imitation targets guide the policy to track 3D human motion while discovering object poses consistent with 2D observations.

\subsection{Generalizability of Universal Controllers}
\label{subsec:universal_generalization}

\begin{figure}[t]\centering
\includegraphics[width=\linewidth]{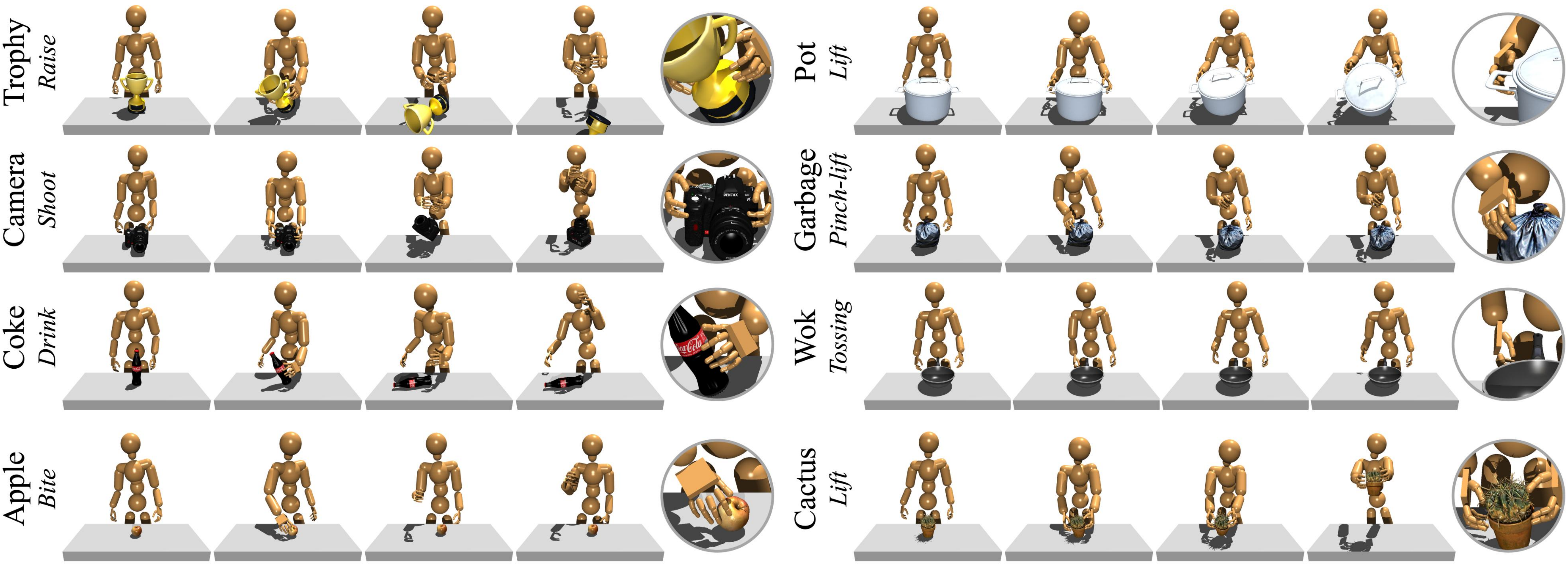}
\caption{\textbf{Limited Generalizability of Universal Controller.} Due to the scarcity of 4D HOI data, an existing universal controller fails to imitate novel HOI motions, such as those reconstructed from videos.
}
\label{fig:universal_failure}
\vspace{-10pt}
\end{figure}

As high-quality 4D HOI data remain scarce, existing datasets provide limited coverage of object geometries, grasps, and manipulation patterns.
While universal controllers such as InterMimic~\citep{InterMimic} imitate diverse HOI motions with a single policy, their generalizability still depends on the coverage of the training demonstrations.
For this evaluation, we use InterMimic's pre-trained checkpoint to imitate precise, physically feasible 4D HOI motions obtained from simulated rollouts of policies trained with DeVI.
These reference motions include both humanoid and object states over time.
As shown in Fig.~\ref{fig:universal_failure}, InterMimic still fails to establish or maintain the hand-object contacts required for novel dexterous interactions (\eg, drinking from a coke bottle and shooting with a camera).
These results show that even precise, physically feasible 4D references do not guarantee successful imitation of novel video-based motion plans by an existing universal controller.
We address this limitation by learning a separate policy for each video with our hybrid imitation targets, adapting the controller to the specific interaction.

\subsection{HOI MoCap Imitation}
\label{subsec:additional_mocap_imitation}

\begin{figure}[t]\centering
\includegraphics[width=1.0\linewidth]{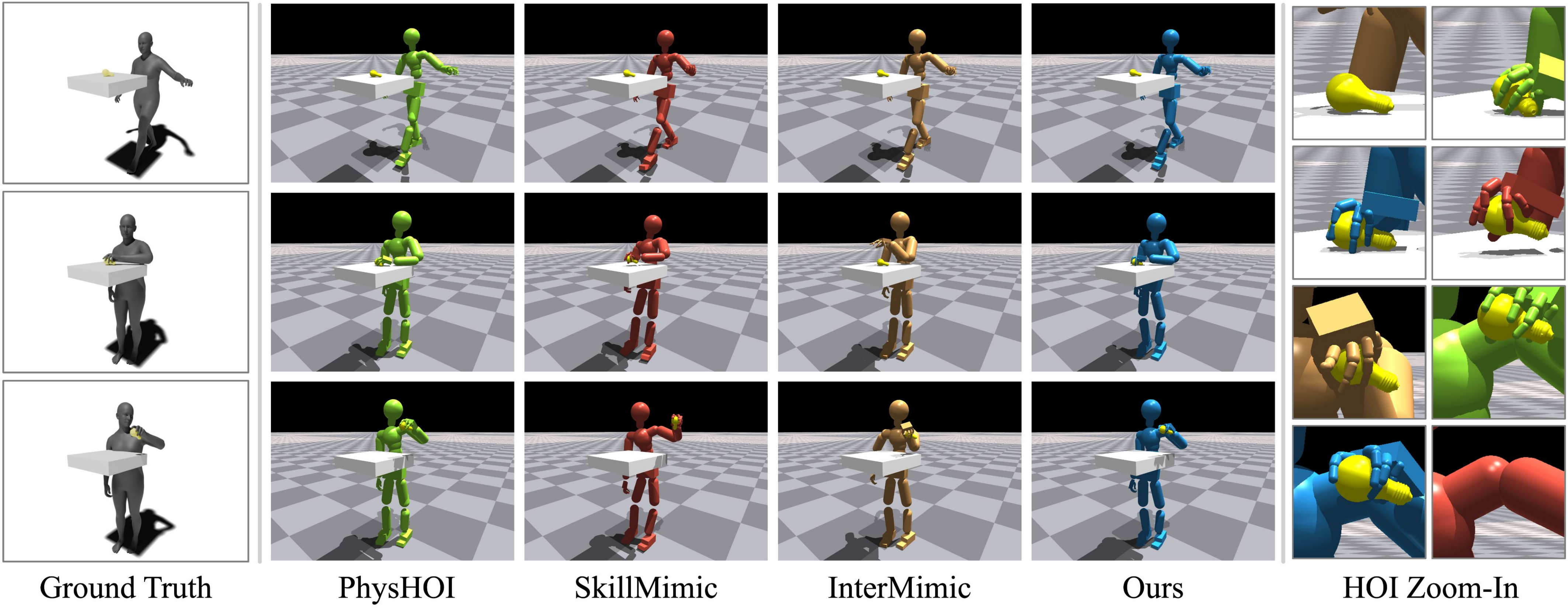}
\caption{\textbf{Qualitative Comparison with MoCap Imitation Baselines.} Even without 6D object poses, DeVI tracks both the human and the object better than baselines using a 6D object pose reward.
}
\label{fig:qual_comparison}
\end{figure}

\noindent\textbf{Qualitative Comparison with Baselines.}
We additionally showcase the qualitative results of DeVI and the baselines in Fig.~\ref{fig:qual_comparison}.
Although DeVI only leverages 2D object trajectories (without full 6D poses), we show results comparable to those of the baselines.
Specifically, DeVI better follows human poses compared to SkillMimic~\citep{SkillMimic} and InterMimic~\citep{InterMimic}, and better follows object poses compared to PhysHOI~\citep{PhysHOI}.
We demonstrate that our hybrid tracking reward effectively imitates MoCap data even without leveraging precise 6D object poses.

\begin{figure*}[t]\centering
\includegraphics[width=\linewidth]{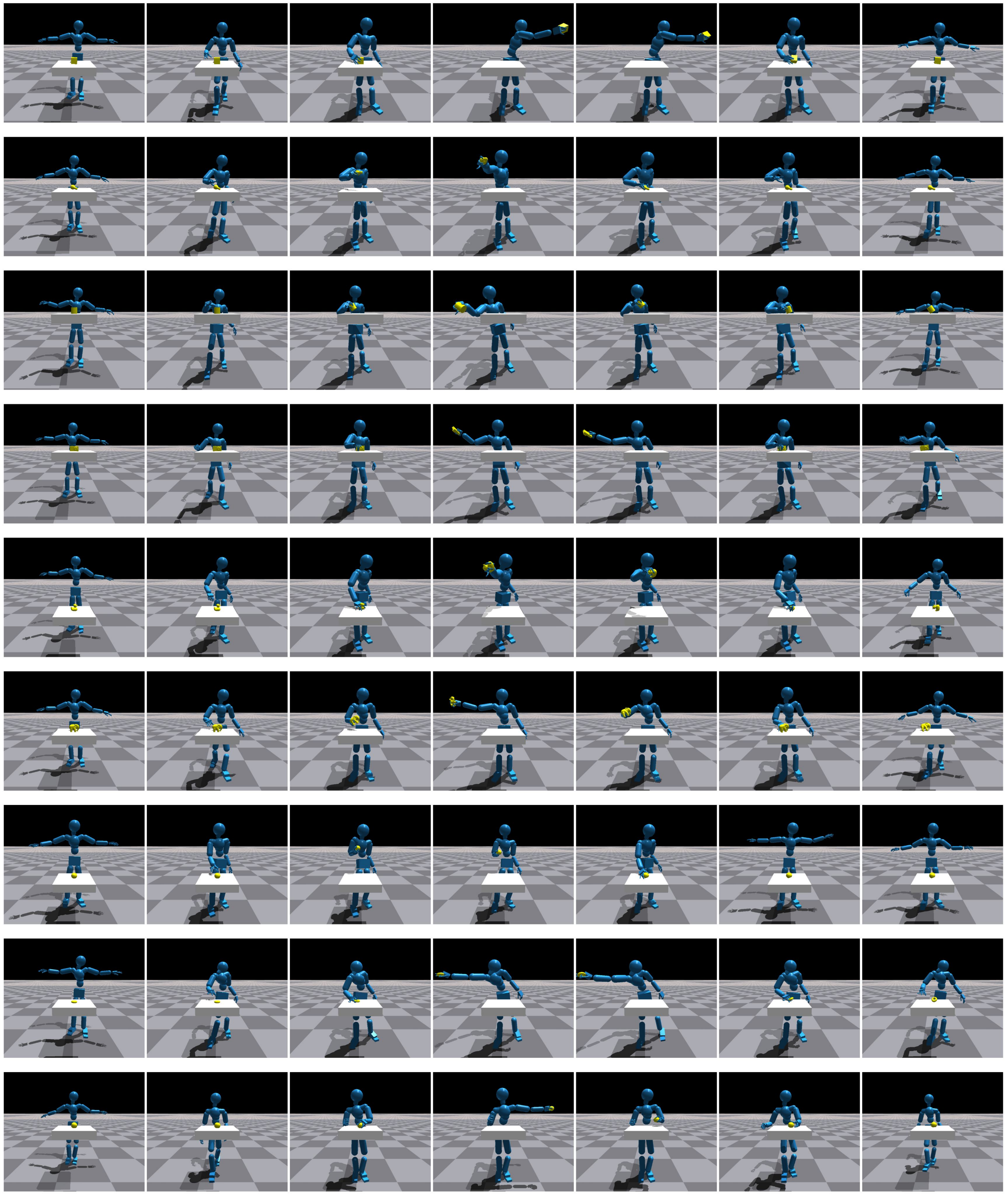}
\captionof{figure}{\textbf{Additional Results of DeVI on GRAB Dataset.} 
}
\label{fig:qual_grab_01}
\end{figure*}

\begin{figure*}[t]\centering
\includegraphics[width=\linewidth]{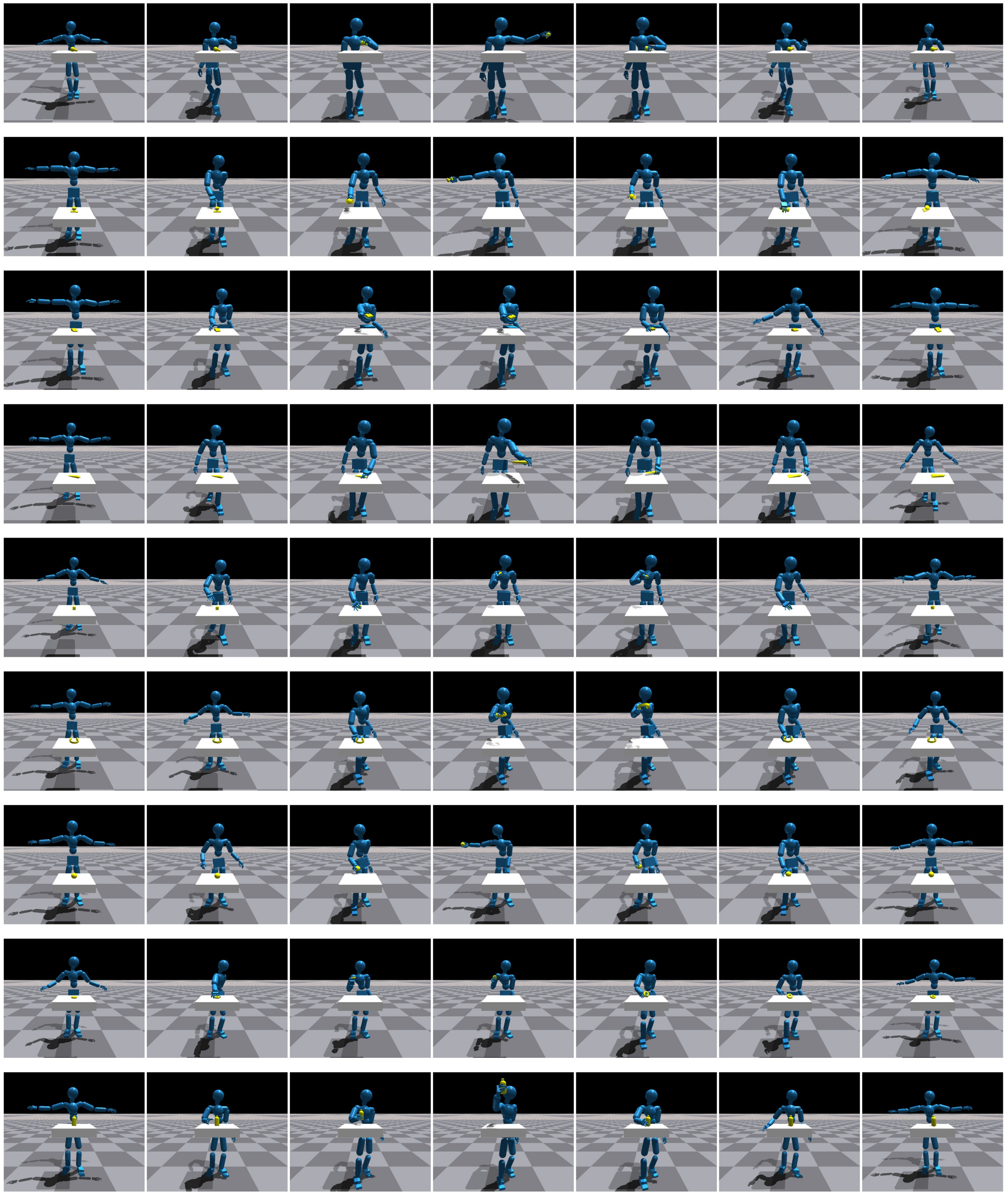}
\captionof{figure}{\textbf{Additional Results of DeVI on GRAB Dataset.} 
}
\label{fig:qual_grab_02}
\end{figure*}

\noindent\textbf{DeVI on GRAB Dataset.} We showcase additional qualitative results in Fig.~\ref{fig:qual_grab_01} and Fig.~\ref{fig:qual_grab_02}.
The figures show the imitation results of our humanoid control policy network on the GRAB~\citep{GRAB} dataset.
As we use only 2D trajectories for learning the humanoid control policy, we project the 3D object vertices into a virtual camera view and use the resulting 2D trajectories as a reference imitation target.
The results show that our method successfully imitates HOI motion using relatively sparse signals, even for HOI scenarios that we do not generate via a video diffusion model.

\subsection{Generalization and Applications}
\label{subsec:application_details}

\begin{figure}[t]\centering
\includegraphics[width=\linewidth]{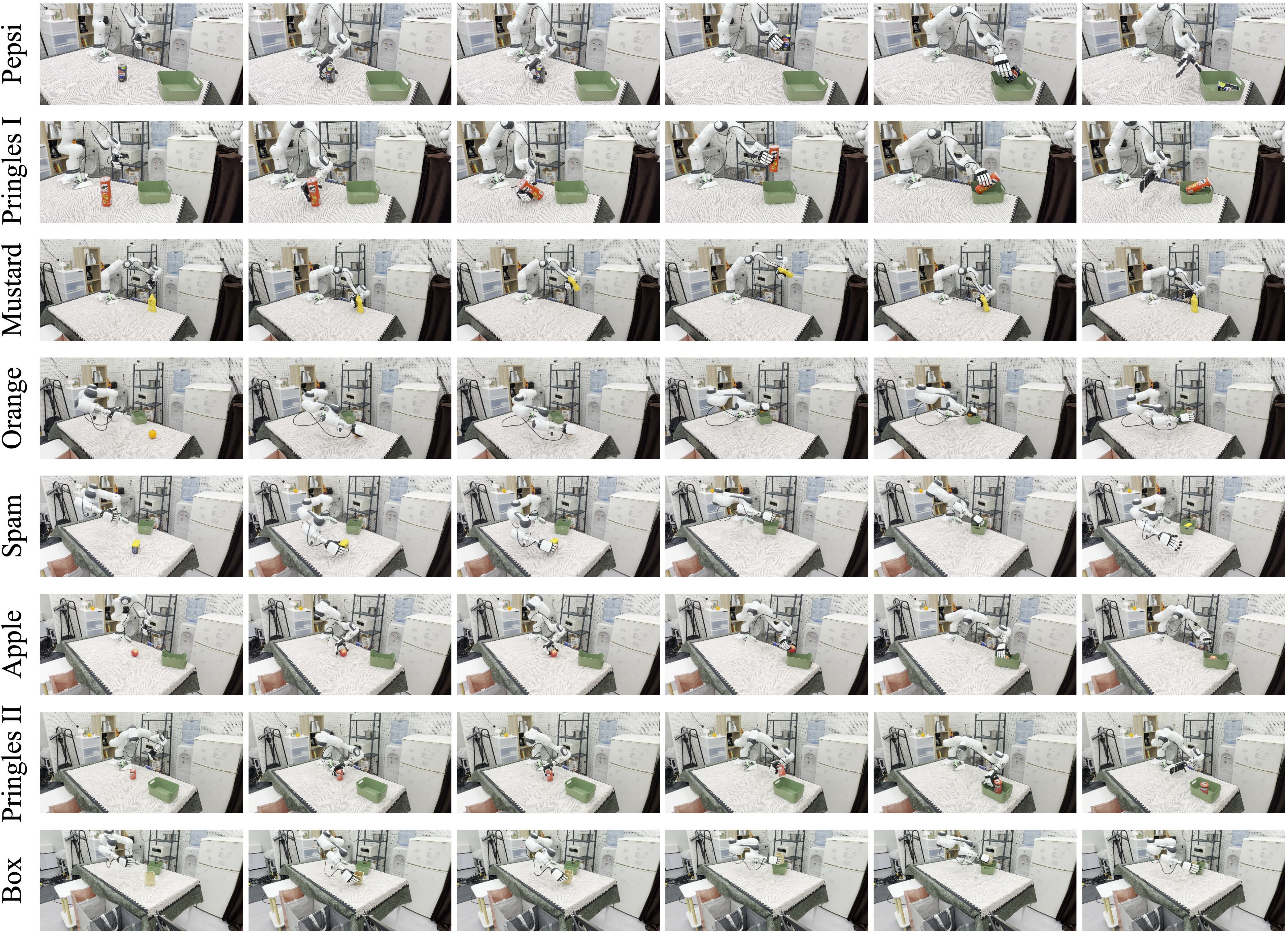}
\caption{\textbf{Real-World Dexterous Manipulation.} Our hybrid imitation target serves as an effective cost for MPC (Model Predictive Control) to optimize a trajectory that can be executed in the real world using a robotic arm equipped with a dexterous hand.
}
\label{fig:realworld}
\vspace{-5pt}
\end{figure}

\noindent\textbf{Real-World Manipulation.}
While our main evaluation focuses on physics-based imitation in simulation, we additionally show the applicability of our hybrid imitation target and Visual HOI Alignment in a real-world environment using a robotic arm equipped with a dexterous hand.
Specifically, we generate egocentric videos and extract hybrid imitation targets from them.
Using these targets, we define the cost for model predictive control (MPC) and find a feasible robotic hand trajectory that minimizes this cost in a physics simulator.
We then execute the optimized trajectory on the real robot in an open-loop manner by supplying the trajectory as PD targets.
Following Dexonomy~\citep{Dexonomy}, we additionally tighten the grasp at contact onset to hold the object firmly.
As shown in Fig.~\ref{fig:realworld}, the robot performs pick-and-place motions with everyday objects.
These examples demonstrate coordinated arm and finger motion for grasping, transporting, and releasing objects with different geometries in a physical environment.

\begin{figure}[t]\centering
\includegraphics[width=\linewidth]{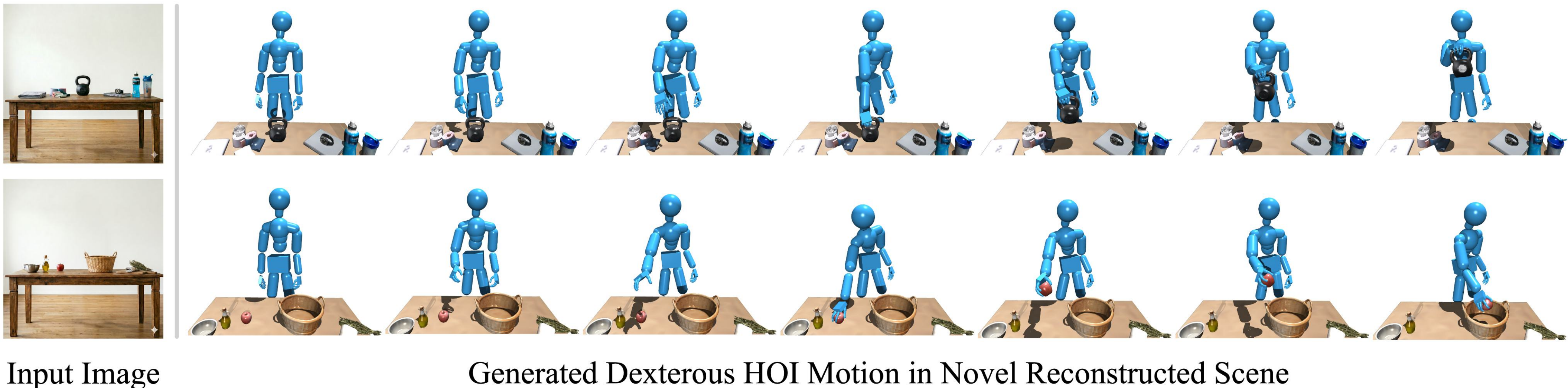}
\caption{\textbf{Generalization of DeVI beyond controlled environments.} DeVI generalizes to novel scenes generated via Image-to-3D.
}
\label{fig:scene_recon}
\vspace{-10pt}
\end{figure}

\noindent\textbf{Generalization beyond Controlled Environments.} As DeVI leverages synthetic videos instead of 3D motion-capture data, it generalizes to novel scenes, even from a single scene image.
Specifically, we generate a reference scene image from a text prompt using Gemini~\citep{gemini} and use SimuScene~\citep{SimuScene} to construct the simulation-ready scene with various assets.
We replace the initial scene with the reconstructed scene to generate physics-based character motion interacting with the objects in the scene.
As shown in Fig.~\ref{fig:scene_recon}, HOI motions within a reconstructed scene can be generated inside a physics simulator, demonstrating our pipeline is broadly applicable across a wide range of environments.

\begin{figure}[t]\centering
\includegraphics[width=\linewidth]{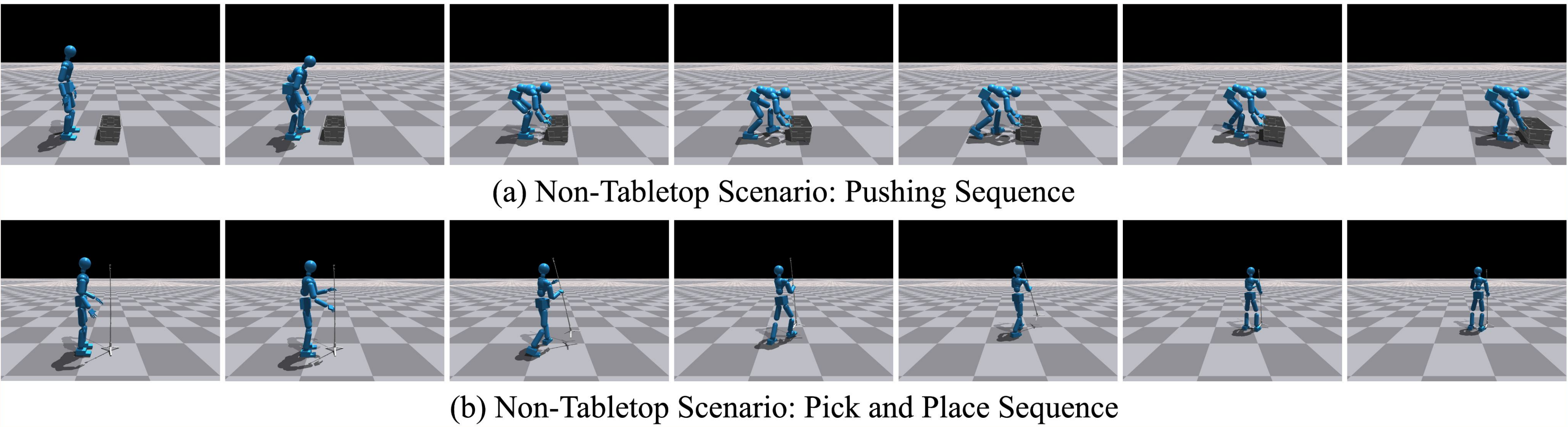}
\caption{\textbf{Non-Tabletop Scenarios.} Our hybrid imitation targets are not limited to tabletop scenarios and can be applied to non-tabletop motions such as (a) pushing and (b) pick and place.
}
\label{fig:non_tabletop}
\vspace{-5pt}
\end{figure}

\noindent\textbf{Non-tabletop Scenarios.} While we intentionally scoped our evaluation to tabletop setups to focus on dexterous HOI imitation from synthetic video without relying on 3D MoCap signals, our hybrid tracking reward is not limited to tabletop scenarios. Fig.~\ref{fig:non_tabletop} shows the results of DeVI trained on the FullBodyManip~\citep{OMOMO} dataset. As shown in Fig.~\ref{fig:non_tabletop}, DeVI and the hybrid imitation rewards extend beyond tabletop scenarios and can also be applied to non-tabletop environments.

\noindent\textbf{Articulated Object Manipulation.}
For the articulated object manipulation in Fig.~\ref{fig:application}, we initialize a NIMBLE~\citep{NIMBLE} hand mesh in the 3D scene and render the initial image from an egocentric viewpoint.
We manually frame the hand mesh so that the cut surface at the wrist lies below the bottom image boundary, while the hand remains visible.
We empirically find that this initialization helps the video generator account for the hand scale when synthesizing the interaction.
Following ManipTrans~\citep{ManipTrans}, we retarget the hand trajectory estimated from the generated video to the simulated Inspire Hand RH56DFTP, providing the hand reference for policy learning.

\begin{figure}[t]\centering
\includegraphics[width=\linewidth]{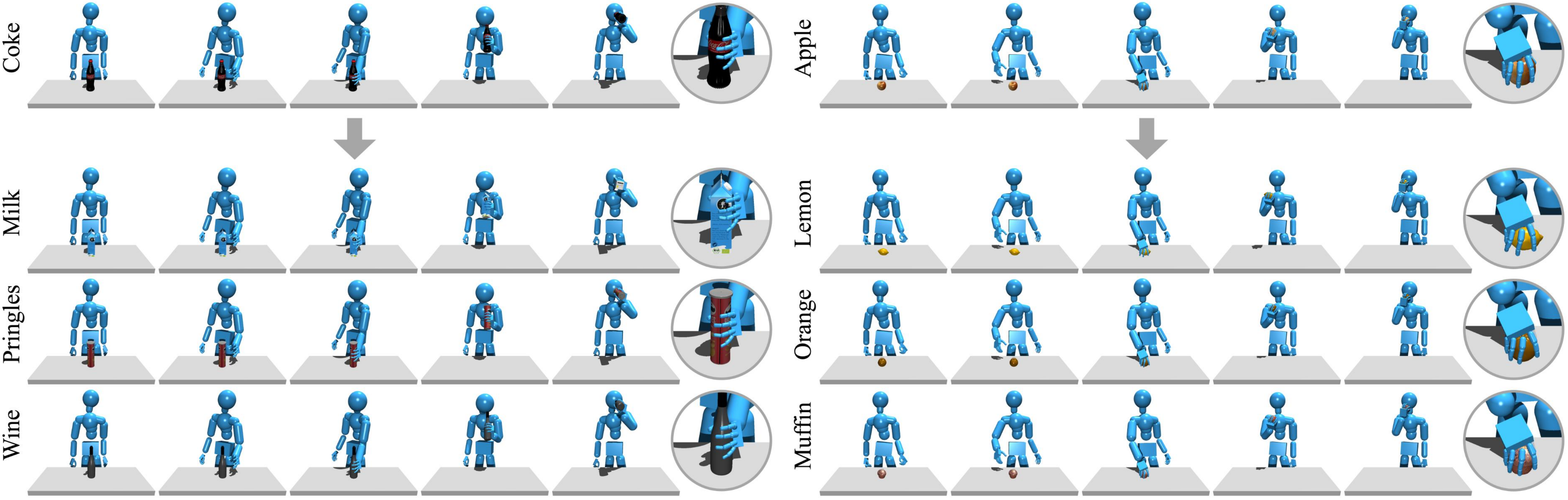}
\caption{\textbf{Zero-Shot Cross-Object Policy Transfer.} Our policy trained to imitate the video can be transferred to an unseen object with similar geometry.
}
\label{fig:transfer}
\vspace{-5pt}
\end{figure}

\noindent\textbf{Cross-Object Policy Transfer.}
As shown in Fig.~\ref{fig:transfer}, our trained policy can be transferred to unseen object, allowing it to generate the same motion while manipulating different objects.
Specifically, we replace the source object with an object of similar geometry and reuse the trained policy without fine-tuning.
The objects have similarly located origins at the bottom of their meshes, and we place these origins at the same position on the table.

\subsection{Stage-wise Evaluation of DeVI}
\label{subsec:stagewise}

Our pipeline chains video generation, human reconstruction, object tracking, and policy learning, so errors in one stage may propagate to the next.
As generated videos lack paired 3D ground truth, we report a stage-specific metric with an explicit success threshold for each stage on the same 15 generated videos.
Due to resource constraints, we restrict the comparisons in Tab.~\ref{tab:quant_video}, Tab.~\ref{tab:visual_hoi_ablation}, and Tab.~\ref{tab:quant_4d_hoi} to a shared 10-video subset of these 15 videos.
The subset is fixed before running the baseline and ablation experiments, supporting fair comparisons across methods.
As shown in Tab.~\ref{tab:stage_video}, Tab.~\ref{tab:stage_human}, Tab.~\ref{tab:stage_track}, and Tab.~\ref{tab:stage_policy}, the stage-wise evaluation of DeVI covers all 15 videos.

\noindent\textbf{Video Generation.}
We conduct a user study in which 71 participants rate each generated video from 1 to 5 for naturalness, physical validity, text control, and overall quality.
We regard a video as successful when all four mean scores reach 4.0.
Under this criterion, 13 of 15 videos succeed.
The failures of Apple and Coke reflect missing object state changes, as the eaten apple shows no bite mark and the Coke bottle stays capped.
These cases limit visual realism but not rigid-object HOI imitation, which is our intended use.

\noindent\textbf{Human Reconstruction.}
We report the 2D reprojection error of the reconstructed SMPL-X joints and the average finger-to-object distance at annotated contact frames.
With failure defined as a hand MPJPE above 10 px or a human-object distance above 50 mm, all 15 videos succeed.
This shows that our Visual HOI Alignment produces human motion consistent with the observed 2D joints and with the 3D object.

\noindent\textbf{Object Tracking.}
We track each point forward and then backward and measure the displacement from its start over points with visibility above $0.1$, which reports drift without ground truth.
We also report $R_{v>\tau}$, the percentage of points retained at visibility thresholds $\tau=0.1$ and $\tau=0.75$.
With failure defined as a mean forward-backward error above 10 px, 13 of 15 videos succeed.

\noindent\textbf{Policy Learning.}
We compare the projected vertices of the simulated object with the 2D tracks, compare the simulated and reference joints, and measure penetration as the bidirectional depth from the human and object SDFs.
With failure defined as an object 2D error above 20 px or a human MPJPE above 50 mm, policy learning succeeds on 14 of 15 videos.
Book is the only failure, which we attribute to the difficulty of inserting fingers beneath its flat underside.
Notably, Camera and Hammer fail at the tracking stage yet still yield successful policies, since the reconstructed human motion and the contact reward jointly constrain the interaction and let the policy settle on an object trajectory that best agrees with the noisy tracks.
This recovery indicates that a failure in one stage does not necessarily propagate to the final policy.

\input{contents/07_stagewise_upper}
\input{contents/08_stagewise_lower}

\section{Limitations and Future Work}
\label{sec:limitation}

\subsection{Perspective Artifacts in Video Diffusion}
While we render a checkered grid floor alongside the scene to provide perspective cues to the video diffusion model, the model often does not produce results with perfect perspective.
For example, when a human moves their hand toward the camera, the hand may appear relatively larger or smaller than it should.
Such perspective artifacts introduce depth-direction errors in our Visual HOI Alignment.
While our HOI loss (based on Chamfer distance) minimizes these errors near the frame where contact starts, errors in the remaining frames may reduce the naturalness of the reconstructed motion.
In particular, this error becomes significant in interactions that require precise target placement (\eg, putting a baseball into a small cup).
As a future direction, we can consider leveraging a multi-view video diffusion model to reduce this depth error.

\subsection{Visibility Loss in 2D Object Tracking}
While our 2D object tracks avoid the unreliable reconstruction of 3D object motion, the tracks are defined on the object vertices visible in the first frame, so the target weakens once those vertices leave the view.
This happens most often under a rapid change of object orientation, where the tracked surface rotates away from the camera within a few frames and the tracker either loses the points or drifts.
As shown in Tab.~\ref{tab:stage_track}, Camera shows the largest forward-backward error among all videos, as the person lifts the camera and turns it toward the face.
Four videos retain no point at the stricter visibility threshold, leaving the object target with little information during the corresponding interval.
Although we filter heavily occluded vertices using the occlusion mask of the tracker, filtering does not recover a surface that rotates out of view.
The same limitation applies to in-hand reorientation and to an object that flips completely, where no vertex from the first frame stays visible.
While the policy still succeeds for Camera in Tab.~\ref{tab:stage_policy} because the human reference and the contact reward constrain the interaction, the object target contributes little during such an interval.

\subsection{Unified Policy Learning}
While our policy learning is fully automatic and applies the same rewards and training setup to every video without per-video tuning, DeVI still trains a separate policy for each generated video, which makes policy learning the most expensive stage of our pipeline.
Learning a single policy conditioned directly on 2D object tracks remains unexplored, so our scope is limited to obtaining diverse functional manipulations rather than a controller that covers every interaction.
As a future direction, conditioning a unified controller on the hybrid imitation target would allow a new video to be imitated through inference rather than training.

\newpage

%% file: contents/09_quant_4d_hoi.tex
\begin{table}[t]
\centering
\small
\setlength{\tabcolsep}{3.5pt}
\renewcommand{\arraystretch}{1.1}
\begin{tabular*}{\linewidth}{@{\extracolsep{\fill}}lccccccccc@{}}
    \toprule
    \multirow{2}{*}{Methods} &
    \multicolumn{3}{c}{MPJPE (px) $\downarrow$} &
    \multirow{2}{*}{\shortstack{IoU\\(\%) $\uparrow$}} &
    \multirow{2}{*}{\shortstack{Precision\\(\%) $\uparrow$}} &
    \multirow{2}{*}{\shortstack{Recall\\(\%) $\uparrow$}} &
    \multirow{2}{*}{\shortstack{F1\\(\%) $\uparrow$}} &
    \multicolumn{2}{c}{Success (\%) $\uparrow$} \\
    \cmidrule(lr){2-4}
    \cmidrule(lr){9-10}
    & Body & Fingers & All & & & & & Frame & Clip \\
    \midrule
    CARI4D
    & 157.66 & 126.29 & 139.57
    & 37.49 & 43.41 & 42.70 & 42.88
    & 5.05 & 0.00 \\
    HOI-PAGE
    & 170.41 & 177.71 & 174.63
    & 42.68 & 48.88 & 48.67 & 48.58
    & 3.67 & 0.00 \\
    \midrule
    DeVI
    & \textbf{34.21} & \textbf{11.86} & \textbf{21.32}
    & \textbf{68.81} & \textbf{80.85} & \textbf{77.71} & \textbf{78.36}
    & \textbf{83.33} & \textbf{90.00} \\
    \bottomrule
\end{tabular*}
\caption{\textbf{Quantitative Comparison with 4D HOI Reconstruction.} We compare the imitation policies trained with references from CARI4D and HOI-PAGE against DeVI. Following the evaluation in Tab.~\ref{tab:visual_hoi_ablation}, we measure human MPJPE in image space using video keypoints as pseudo ground truth.}
\label{tab:quant_4d_hoi}
\end{table}

%% file: contents/07_stagewise_upper.tex
\begin{table}[h]
\centering
\begin{minipage}[t]{0.48\linewidth}
    \vspace{0pt}
    \centering
    \scriptsize
    \setlength{\tabcolsep}{2.0pt}
    \begin{tabular*}{\linewidth}{@{\extracolsep{\fill}}lccccc@{}}
        \toprule
        \multirow{2}{*}{\raisebox{-2.0pt}{Object}} &
        \multicolumn{4}{c}{User Study $\uparrow$} &
        \multirow{2}{*}{\raisebox{-2.0pt}{S/F}} \\
        \cmidrule(lr){2-5}
        & Nat. & Phys. & Text & Overall & \\
        \midrule
        Apple     & 3.75 & 3.73 & 3.90 & 3.70 & F \\
        Book      & 4.17 & 4.46 & 4.48 & 4.28 & S \\
        Cactus    & 4.23 & 4.37 & 4.51 & 4.28 & S \\
        Camera    & 4.15 & 4.32 & 4.49 & 4.23 & S \\
        Coke      & 3.69 & 3.44 & 4.06 & 3.61 & F \\
        Cookpot   & 4.03 & 4.08 & 4.34 & 4.06 & S \\
        Garbage   & 4.28 & 4.35 & 4.49 & 4.32 & S \\
        Hammer     & 4.07 & 4.31 & 4.30 & 4.17 & S \\
        Mug       & 4.30 & 4.20 & 4.28 & 4.13 & S \\
        Pot       & 4.27 & 4.42 & 4.39 & 4.28 & S \\
        Softdrink & 4.30 & 4.41 & 4.56 & 4.31 & S \\
        Strawhat  & 4.31 & 4.27 & 4.42 & 4.27 & S \\
        Teapot    & 4.39 & 4.51 & 4.41 & 4.37 & S \\
        Trophy    & 4.11 & 4.34 & 4.42 & 4.13 & S \\
        Wok       & 4.13 & 4.23 & 4.38 & 4.14 & S \\
        \midrule
        Mean      & 4.14 & 4.23 & 4.36 & 4.15 & 13/15 \\
        \bottomrule
    \end{tabular*}
    \captionof{table}{\textbf{Video Generation.} Video generation fails when any of the four scores falls below $4.0$.}
    \label{tab:stage_video}
\end{minipage}\hfill
\begin{minipage}[t]{0.48\linewidth}
    \vspace{0pt}
    \centering
    \scriptsize
    \setlength{\tabcolsep}{2.0pt}
    \begin{tabular*}{\linewidth}{@{\extracolsep{\fill}}lccccc@{}}
        \toprule
        \multirow{2}{*}{\raisebox{-2.0pt}{Object}} &
        \multicolumn{3}{c}{MPJPE (px) $\downarrow$} &
        \multirow{2}{*}{\raisebox{-9.0pt}{\shortstack{HOI Distance\\(mm) $\downarrow$}}} &
        \multirow{2}{*}{\raisebox{-2.0pt}{S/F}} \\
        \cmidrule(lr){2-4}
        & Body & Hand & All & & \\
        \midrule
        Apple     & 15.3 & 3.16 & 7.09 & 19.2 & S \\
        Book      & 53.2 & 5.33 & 20.8 & 31.3 & S \\
        Cactus    & 27.0 & 2.98 & 10.7 & 15.8 & S \\
        Camera    & 32.4 & 7.04 & 15.2 & 19.2 & S \\
        Coke      & 29.1 & 3.39 & 11.7 & 10.9 & S \\
        Cookpot   & 29.9 & 3.86 & 12.3 & 16.4 & S \\
        Garbage   & 33.9 & 2.88 & 12.9 & 14.3 & S \\
        Hammer     & 30.9 & 3.65 & 12.4 & 28.0 & S \\
        Mug       & 43.2 & 4.67 & 17.1 & 19.2 & S \\
        Pot       & 34.6 & 5.37 & 14.8 & 27.0 & S \\
        Softdrink & 18.6 & 2.38 & 7.59 & 27.2 & S \\
        Strawhat  & 27.2 & 4.38 & 11.8 & 14.1 & S \\
        Teapot    & 32.7 & 6.19 & 14.8 & 33.9 & S \\
        Trophy    & 32.0 & 4.44 & 13.3 & 45.3 & S \\
        Wok       & 24.1 & 4.78 & 11.0 & 24.4 & S \\
        \midrule
        Mean      & 31.0 & 4.30 & 12.9 & 23.1 & 15/15 \\
        \bottomrule
    \end{tabular*}
    \captionof{table}{\textbf{Human Reconstruction.} 
    Human reconstruction fails when hand MPJPE exceeds 10 px or HOI distance exceeds 50 mm.
    }
    \label{tab:stage_human}
\end{minipage}
\end{table}

%% file: contents/08_stagewise_lower.tex
\begin{table}[h]
\centering
\begin{minipage}[t]{0.48\linewidth}
    \vspace{0pt}
    \centering
    \scriptsize
    \setlength{\tabcolsep}{2.0pt}
    \begin{tabular*}{\linewidth}{@{\extracolsep{\fill}}lcccc@{}}
        \toprule
        \multirow{2}{*}{\raisebox{-2.0pt}{Object}} &
        \multirow{2}{*}{\raisebox{-9.0pt}{\shortstack{FB Err.\\(px) $\downarrow$}}} &
        \multicolumn{2}{c}{Retained (\%) $\uparrow$} &
        \multirow{2}{*}{\raisebox{-2.0pt}{S/F}} \\
        \cmidrule(lr){3-4}
        & & $R_{v>0.1}$ & $R_{v>0.75}$ & \\
        \midrule
        Apple     & 6.75 & 68.7 & 0.00 & S \\
        Book      & 5.71 & 100  & 71.2 & S \\
        Cactus    & 1.95 & 93.9 & 67.3 & S \\
        Camera    & 26.4 & 99.5 & 43.7 & F \\
        Coke      & 4.70 & 100  & 3.10 & S \\
        Cookpot   & 6.65 & 95.6 & 37.2 & S \\
        Garbage   & 3.99 & 89.1 & 45.5 & S \\
        Hammer     & 10.6 & 93.0 & 0.00 & F \\
        Mug       & 4.90 & 82.2 & 0.00 & S \\
        Pot       & 1.00 & 99.6 & 87.0 & S \\
        Softdrink & 5.29 & 25.7 & 7.70 & S \\
        Strawhat  & 8.02 & 88.0 & 0.00 & S \\
        Teapot    & 2.01 & 83.8 & 58.6 & S \\
        Trophy    & 1.10 & 100  & 83.1 & S \\
        Wok       & 3.83 & 45.5 & 22.7 & S \\
        \midrule
        Mean      & 6.19 & 84.3 & 35.1 & 13/15 \\
        \bottomrule
    \end{tabular*}
    \captionof{table}{\textbf{Object Tracking.} 
    Object tracking fails when the forward-backward error exceeds 10 px.
    }
    \label{tab:stage_track}
\end{minipage}\hfill
\begin{minipage}[t]{0.48\linewidth}
    \vspace{0pt}
    \centering
    \scriptsize
    \setlength{\tabcolsep}{2.0pt}
    \begin{tabular*}{\linewidth}{@{\extracolsep{\fill}}lcccc@{}}
        \toprule
        \multirow{2}{*}{\raisebox{-2.0pt}{Object}} &
        \multicolumn{2}{c}{Tracking Err. $\downarrow$} &
        \multirow{2}{*}{\raisebox{-9.0pt}{\shortstack{Penetration \\(mm) $\downarrow$}}} &
        \multirow{2}{*}{\raisebox{-2.0pt}{S/F}} \\
        \cmidrule(lr){2-3}
        & Object (px) & Human (mm) & & \\
        \midrule
        Apple     & 9.36 & 26.4 & 1.42  & S \\
        Book      & 356  & 612  & 0.545 & F \\
        Cactus    & 5.70 & 39.8 & 3.90  & S \\
        Camera    & 4.83 & 27.9 & 7.09  & S \\
        Coke      & 3.72 & 25.9 & 2.12  & S \\
        Cookpot   & 10.6 & 19.8 & 1.70  & S \\
        Garbage   & 6.63 & 29.5 & 2.88  & S \\
        Hammer     & 3.84 & 26.2 & 1.77  & S \\
        Mug       & 4.95 & 27.3 & 1.65  & S \\
        Pot       & 3.36 & 34.9 & 3.37  & S \\
        Softdrink & 6.66 & 42.4 & 0.860 & S \\
        Strawhat  & 15.6 & 42.8 & 2.32  & S \\
        Teapot    & 3.55 & 27.3 & 0.950 & S \\
        Trophy    & 3.29 & 31.8 & 3.16  & S \\
        Wok       & 13.6 & 34.4 & 1.90  & S \\
        \midrule
        Mean      & 30.1 & 69.9 & 2.38  & 14/15 \\
        \bottomrule
    \end{tabular*}
    \captionof{table}{\textbf{Policy Learning.} 
    Policy learning fails when object tracking error exceeds 20 px or human tracking error exceeds 50 mm.
    }
    \label{tab:stage_policy}
\end{minipage}
\end{table}